\pdfoutput=1
\documentclass[letterpaper]{article}
\usepackage[preprint]{kosmo}

\usepackage{multirow}
\usepackage{xspace}
\usepackage{float}
\usepackage{needspace}

\newcommand{\method}{\textsc{MVLoMa}\xspace}
\newcommand{\da}{DA3\xspace}

\title{MVLoMa: Multi-View Local Feature Matching with Any Geometry
Foundation Model}

\author{
  Xinyang Lu\textsuperscript{1,2},
  Shengzhe Wei\textsuperscript{1},
  Sheng Tang\textsuperscript{1},
  Weiqi Zhang\textsuperscript{3},
  Juntong Fang\textsuperscript{3},
  Zesong Li\textsuperscript{1}
}

\affiliations{
  \textsuperscript{1}Kosmo Research\quad
  \textsuperscript{2}School of Astronautics, Harbin Institute of Technology\\
  \textsuperscript{3}School of Software, Tsinghua University
}

\correspondingauthor{}

\projectpage{\url{https://kosmoresearch.github.io/MVLoMa/}}

\keywords{local feature matching, multi-view geometry, star graph,
geometry foundation models, relative pose estimation}

\kosmoabstract{
Local feature matching is typically formulated for image pairs, although
structure-from-motion and visual localization commonly match one reference
image against many retrieved or covisible neighbours. Applying a pairwise
matcher independently to these centre--neighbour edges duplicates
reference-side computation and prevents the matcher from exploiting
complementary multi-view geometric evidence from the remaining observations.
To address these limitations, we introduce MVLoMa, a sparse star-graph
matcher that extends LoMa from image pairs to multiple views. MVLoMa first
uses a frozen multi-view geometry model to jointly describe all images in a
star, then injects the resulting cross-view geometric context into local
descriptors. It then matches the centre image to all neighbours in a single
forward pass through per-view self-attention and bidirectional
centre--neighbour cross-attention. For supervision, we derive positive
correspondences only from depth-valid, mutually visible dense warps and
explicitly penalize high-confidence matches that violate epipolar geometry.
On 25-view star graphs from HyperSim, ScanNet++, Mip-NeRF 360, and Tanks and
Temples, MVLoMa obtains the best macro-average relative-pose AUC among the
compared methods at all three thresholds (58.65/85.41/90.99). Its matcher is
$2.20\times$ faster than sequential evaluation of the 24 centre--neighbour
edges. The same design supports four tested frozen geometry
encoders---DA3, $\pi^3$, VGGT, and VGGT-$\Omega$---with DA3-Giant performing
best on all reported encoder-ablation metrics. Controlled ablations further
isolate the effects of geometry-aware descriptors, visibility filtering,
geometric negative supervision, and joint multi-view inference.
}

\begin{document}

\maketitle

\section{Introduction}
\label{sec:intro}

Reliable correspondence is the interface through which images become
geometric constraints. It underlies camera pose estimation, mapping, visual
localization, and structure-from-motion (SfM). Learned sparse
matchers~\citep{sarlin2020superglue,lindenberger2023lightglue,nordstrom2026loma}
combine local descriptors with contextual assignment, while detector-free
and dense methods predict a broader set of
correspondences~\citep{sun2021loftr,edstedt2024roma,edstedt2026romav2}.
This progress has made two-view pose estimation substantially more robust.
Yet it also leaves a mismatch between the unit of computation used by a
matcher---an image pair---and the unit of evidence available to downstream
geometry---a group of mutually overlapping images.

Consider the ubiquitous star graph: a reference image and its retrieved
neighbours. This topology occurs when SfM builds local subproblems, when
localization matches a query to a covisible map neighbourhood, and when a
keyframe associates observations over time. A pairwise matcher is invoked
once for every reference--neighbour edge. It consequently recomputes the
reference descriptor and reference-side matcher branch, and it must decide
each correspondence before seeing supporting or conflicting evidence from
the remaining neighbours. The limitation is not merely redundant
computation. Repetitive texture, partial occlusion, and a weak baseline can
make a match ambiguous in one pair while a third image resolves the
ambiguity.

Existing multi-view matchers move contextualization from isolated pairs to
collaborative groups or tracks~\citep{zhang2025comatcher,lee2026mvroma}.
However, their dense or groupwise outputs are not designed for the repeated
sparse centre--neighbour assignments of a star graph. Simply batching
pairwise matchers preserves that sparse interface but prevents communication
between edges. We therefore seek a matcher that reasons jointly over a star
while retaining explicit pairwise outputs for RANSAC.

There is a second, largely independent source of unused information. Recent
3D foundation models predict pose, depth, and geometry from arbitrary view
collections by exchanging information across image
tokens~\citep{wang2025vggt,wang2025pi3,lin2025da3}. Their spatially aligned
final-layer tokens already encode multi-view geometric context, yet local
matching commonly starts from a separate frozen 2D descriptor backbone.
Reusing these tokens exposes geometry to the matcher without introducing
another frozen feature stream. Keeping the foundation encoder fixed also
avoids optimizing its large parameter set and preserves a reusable geometric
representation.

These observations suggest a simple principle: \emph{jointly match a star
once, and reuse the frozen geometric representation that the star has
already produced}. We instantiate this principle in \method, a multi-view
extension of LoMa~\citep{nordstrom2026loma}. A light trainable descriptor
head replaces LoMa's frozen DINOv2 stream with last-layer tokens from a
frozen 3D foundation encoder. A star-structured matcher then shares the
centre representation across all edges and alternates per-view
self-attention with bidirectional centre--neighbour cross-attention. It
outputs a conventional score matrix for every edge, so its matches can be
consumed directly by standard robust pose estimation.

Joint reasoning alone does not guarantee useful supervision. Sparse
assignment is a classification problem: high-confidence but geometrically
wrong matches can survive mutual matching and disproportionately harm
RANSAC. We therefore construct positives only from depth-valid,
mutually-visible dense warps, and explicitly penalize predicted matches
whose epipolar error exceeds 3\,px. These changes complement the multi-view
architecture rather than replace LoMa's matching and matchability
objectives.

Our experiments follow a 25-view star-graph protocol and evaluate the same
centre--neighbour edges for every method. Under this protocol, \method leads
the macro-average at all three AUC thresholds and improves strict AUC over
LoMa-L on both HyperSim and ScanNet++. A controlled backbone ablation
isolates the effect of geometry-aware features, supervision ablations
evaluate visibility filtering and geometric negatives, and a view-count
study shows that additional views can improve joint multi-view inference.
The results also delineate the scope of the method: dense two-view RoMa
remains stronger on Mip-NeRF 360 at the strictest threshold, while \method
provides a sparse multi-view alternative that operates on all 25 views in
our protocol.

Our contributions are summarized as follows:
\begin{itemize}
  \item We formulate sparse matching over a star so that \method produces
  all centre--neighbour matches in one forward pass rather than evaluating
  each edge independently.
  \item We use frozen features from a multi-view foundation model---\da-Giant
  by default---as geometry-aware descriptor inputs, so matching shares the
  representation already used for multi-view geometry.
  \item We adopt RoMa v2 visibility-filtered ground truth for star-graph
  training~\citep{edstedt2026romav2} and introduce a geometric negative loss
  for high-confidence epipolar violations. Experiments and ablations show
  consistent improvements from these components and from multi-view context.
\end{itemize}

\section{Related Work}

\subsection{Sparse and Dense Feature Matching}

Local matching has evolved from independently comparing hand-designed
features to learned descriptor and assignment pipelines. SuperGlue
contextualizes detected keypoints with graph
attention~\citep{sarlin2020superglue}; LightGlue improves the same family
with adaptive depth and pruning~\citep{lindenberger2023lightglue}. LoFTR
removes the detector by matching coarse image features~\citep{sun2021loftr}.
In parallel, RoMa and RoMa v2 obtain robust dense correspondence from
foundation features and coarse-to-fine
refinement~\citep{edstedt2024roma,edstedt2026romav2}. LoMa shows that a
carefully scaled local-feature recipe remains highly
competitive~\citep{nordstrom2026loma}. These methods differ in sparsity and
representation, but their assignment is fundamentally pairwise. \method
inherits LoMa's local descriptor refinement and dual-softmax interface while
moving contextual reasoning from an image pair to a star graph.

\subsection{Multi-View Correspondence and Tracks}

Multi-view information has long been used after matching to merge pairwise
correspondences into tracks. More recent learned methods use it inside the
matcher. CoMatcher employs groupwise interaction to make keypoint matching
collaborative~\citep{zhang2025comatcher}, whereas MV-RoMa aggregates dense
multi-view correspondence into tracks~\citep{lee2026mvroma}. These
approaches establish that complementary observations can reduce pairwise
ambiguity. Our setting is deliberately more structured: a star has one
centre whose features participate in every output edge. We therefore use a
centre--neighbour communication pattern rather than full
neighbour--neighbour interaction. The resulting model retains a score matrix
and confidence for each individual edge, enabling the usual sparse matching
and RANSAC interface without a post-hoc track snapping stage.

\subsection{Multi-View Geometric Foundation Models}

DUSt3R and MASt3R learn feedforward geometric representations from image
pairs~\citep{wang2024dust3r,leroy2024mast3r}. VGGT, $\pi^3$, and \da extend
the input from pairs to view collections and exchange token information
across views~\citep{wang2025vggt,wang2025pi3,lin2025da3}. These models are
typically used to decode camera and scene geometry, while a separate
matching system constructs explicit correspondences. We instead ask whether
their last-layer, cross-view-fused tokens can serve as a frozen
geometry-aware input to a sparse matcher. This choice is distinct from
directly decoding tracks with a foundation model: it preserves a specialized
assignment head and leaves the geometry encoder unchanged.

\subsection{Matching in Reconstruction Systems}

Optimization-based SfM obtains global geometry by matching image pairs,
building tracks, and geometrically verifying
them~\citep{schoenberger2016colmap,pan2024glomap}. Hybrid systems couple
these explicit constraints with feedforward geometry
priors~\citep{duisterhof2024mast3rsfm,pan2026gluemap}. They motivate star
subgraphs as a scalable local unit, but the correspondence module is still
often assembled from independent pairwise matchers or heterogeneous track
sources. \method addresses this upstream bottleneck; it is a matching
primitive rather than a replacement for global optimization.

\section{Method}
\label{sec:method}

\subsection{Problem Formulation}

\begin{figure*}[t]
  \centering
  \includegraphics[width=0.96\textwidth]{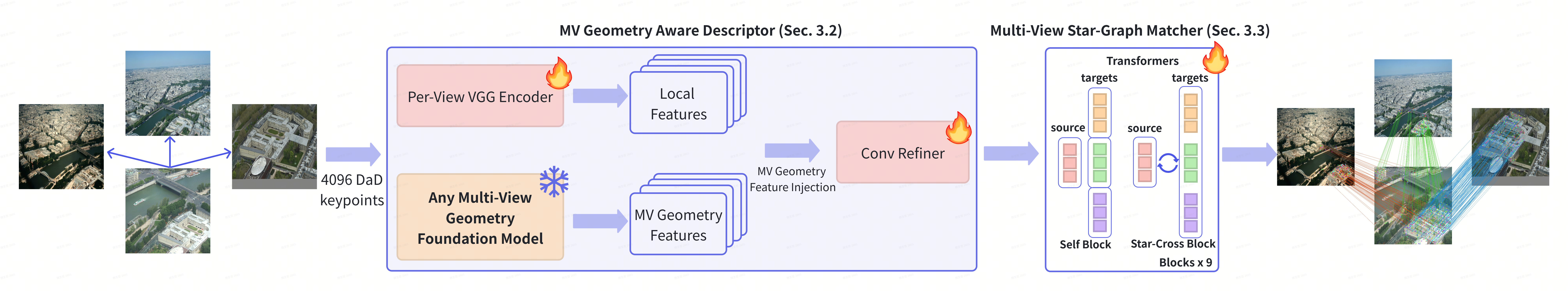}
  \caption{Overview of \method. Given one source (centre) image and
  multiple target (neighbour) images, a trainable per-view VGG branch
  extracts local features, while a frozen multi-view geometry foundation
  model supplies geometry-aware patch features for the same views. A
  trainable convolutional refiner fuses the two streams into descriptors
  sampled at 4096 keypoints per image. The multi-view star-graph matcher
  then jointly processes the centre and all neighbours and outputs sparse
  centre--neighbour correspondences for every edge. Flames denote trainable
  modules and the snowflake denotes the frozen foundation model.}
  \label{fig:architecture}
\end{figure*}

Given a star $\mathcal{S}_l=\{I_l,I_1,\ldots,I_{N-1}\}$ with centre image
$I_l$, our goal is to simultaneously predict the correspondence sets
$\{M(l,j)\}_{j=1}^{N-1}$. Each $M(l,j)$ contains sparse centre--neighbour
matches. A pairwise matcher instead makes $N-1$ independent calls, each with
a duplicate centre descriptor and centre-side attention computation. More
importantly, it commits to an edge without observations from the other
edges. \method is designed around the opposite data flow: frozen foundation
features first encode the complete star, a shared centre representation then
communicates with all neighbours, and conventional pairwise assignment is
recovered only at the output (Figure~\ref{fig:architecture}). The following
sections describe these three stages and their geometric supervision.

\subsection{Geometry-Aware Multi-View Descriptors}
\label{sec:desc}

The first design question is where multi-view context should enter a local
matcher. Recomputing independent 2D descriptors for every edge would
preserve LoMa's pairwise bottleneck. Conversely, fine-tuning a large
geometry encoder solely for matching is costly and can disturb its
general-purpose prior. We instead reuse the representation already available
after a foundation model has processed the star.

Multi-view geometry models tokenize each image with a ViT and interleave
within-view self-attention with cross-view attention. Let
$\mathbf{F}_i\in\mathbb{R}^{h\times w\times d_f}$ be the final
alternating-attention feature map for $I_i$. Its tokens are spatially
aligned and include information from the other star views. The original LoMa
DeDoDe-G descriptor~\citep{nordstrom2026loma} fuses a trainable VGG stream
and convolutional refiners with frozen DINOv2-L
features~\citep{oquab2024dinov2}. \method keeps this trainable structure,
replacing only the frozen DINOv2-L stream with frozen foundation features:
\begin{equation}
  \mathbf{D}_i=g_\psi\big(\mathrm{VGG}(I_i),\mathbf{F}_i\big),
  \quad \mathbf{D}_i\in\mathbb{R}^{h\times w\times d},
  \label{eq:desc}
\end{equation}
for $i\in\{l,1,\ldots,N-1\}$. Here $g_\psi$ comprises the Conv Refiner and
MLP projection, which also maps encoder-specific feature dimensions to a
common descriptor dimension. LoMa's DaD keypoints are detected and
descriptors are sampled from $\mathbf{D}_i$. Thus multi-view context enters
the descriptor field before assignment, while the subsequent matcher can
focus on selecting sparse correspondences. The same star-graph encoder pass
also produces the geometry features used by its original pose/depth heads;
\method introduces no separate frozen descriptor backbone. Compatibility
requires a frozen encoder to expose spatially aligned per-image tokens after
multi-view fusion. We use DA3-Giant by default and evaluate $\pi^3$, VGGT,
and VGGT-$\Omega$~\citep{wang2025pi3,wang2025vggt,wang2026vggtomega} in
Section~\ref{sec:experiments}.

\subsection{Star-Graph Attention and Assignment}
\label{sec:attention}

The descriptor fields are geometry-aware, but matching still benefits from
task-specific communication at the selected keypoints. Running LoMa on each
edge would retain independent centre states. Full all-to-all attention is
also unnecessary for a star: the outputs only connect the centre with a
neighbour, and allowing every neighbour pair to interact increases the
attention workload without providing a direct output edge. We therefore use
the star topology itself as the attention pattern
(Figure~\ref{fig:architecture}).

Starting from sampled descriptors, we stack $L$ matcher layers. Each layer
first forms an intermediate state by applying self-attention independently
within every view, with rotary position encoding:
\begin{equation}
  \mathbf{H}^{(\ell)}_i =
  \mathbf{D}^{(\ell-1)}_i+
  \mathrm{SelfAttn}(\mathbf{D}^{(\ell-1)}_i).
  \label{eq:self}
\end{equation}
We then concatenate neighbour descriptors, writing
$\mathbf{X}_{1:N-1}=[\mathbf{X}_1,\ldots,\mathbf{X}_{N-1}]$ for the stacked
neighbour tokens. The centre queries all neighbours, after which all
neighbours query the updated centre:
\begin{align}
  \mathbf{D}^{(\ell)}_l &= \mathbf{H}^{(\ell)}_l+
  \mathrm{CrossAttn}\Big(\mathbf{H}^{(\ell)}_l;
  \mathbf{H}^{(\ell)}_{1:N-1}\Big),
  \label{eq:crosscentre}\\
  \mathbf{D}^{(\ell)}_{1:N-1}
  &=\mathbf{H}^{(\ell)}_{1:N-1}
  +\mathrm{CrossAttn}\Big(\mathbf{H}^{(\ell)}_{1:N-1};
  \mathbf{D}^{(\ell)}_l\Big).
  \label{eq:crossneighbour}
\end{align}
Neighbours never attend to one another and no view embedding is used;
cross-attention has no positional encoding. The centre can nevertheless
aggregate evidence from all observations before assignment. A point
ambiguous in one pair can consequently be supported or contradicted by a
third view, while each edge retains its own sparse output.

For each neighbour $j$, we form
$S_{pq}^{lj}=\langle\mathbf{d}^l_p,\mathbf{d}^j_q\rangle$ and calculate a
pairwise dual-softmax confidence:
\begin{equation}
  P_{pq}^{lj}=
  \mathrm{softmax}_p(\tau^{-1}S_{pq}^{lj})\odot
  \mathrm{softmax}_q(\tau^{-1}S_{pq}^{lj}).
  \label{eq:dualsoftmax}
\end{equation}
Normalization is restricted to the current image pair; neighbours therefore
compete only for correspondences on their own edge. Mutually maximal entries
above the matching threshold $\mu$ form $M(l,j)$. Since the centre
descriptors have been shared across all $j$, the same centre keypoint can
consistently associate to several neighbours, inducing a candidate star
track without a separate track-snapping operation.

\subsection{Visibility-Filtered Ground Truth and Geometric Negatives}
\label{sec:loss}

The attention pattern establishes how evidence is exchanged; supervision
must ensure that this evidence is geometrically meaningful. We obtain
LoMa-style positive correspondences by reprojecting ground-truth depth with
ground-truth pose. A naive dense warp, however, treats occluded,
out-of-image, and invalid-depth pixels as if they had a valid target.
Following RoMa v2~\citep{edstedt2026romav2}, we retain a positive only when
its source depth is valid, its reprojection lies inside the target image,
and the prescribed occlusion and target-depth consistency checks are
satisfied. This filtering removes label noise before it reaches either the
descriptor or the assignment head.

The original dual-softmax matching loss suppresses non-matches implicitly
through row and column competition, and matchability indicates whether a
keypoint has a true correspondence. Neither objective explicitly punishes a
high-confidence false match that survives the final mutual selection. Such
outliers are particularly consequential for sparse geometric estimation. We
therefore mine predictions that violate ground-truth epipolar geometry and
use them as explicit negatives.

For $(I_l,I_j)$, ground-truth relative extrinsics define
$E^{lj}=[t^{lj}]_\times R^{lj}$ and the fundamental matrix at the network
output resolution is $F^{lj}=K_j^{-\top}E^{lj}K_l^{-1}$. For homogeneous
coordinates $\bar{x}^l_p,\bar{x}^j_q$, the point-to-epipolar-line error is
\begin{equation}
  e_{pq}^{lj}=
  \frac{|(\bar{x}^j_q)^\top F^{lj}\bar{x}^l_p|}
  {\|(F^{lj}\bar{x}^l_p)_{1:2}\|_2}.
  \label{eq:epi}
\end{equation}
Let $\mathcal{M}^{lj}$ denote final-layer mutually maximal matches whose
confidence exceeds $\mu$. We define geometric negatives as
\begin{equation}
  \mathcal{N}^{lj}=\big\{(p,q)\in\mathcal{M}^{lj}\mid
  e_{pq}^{lj}>\tau_{\mathrm{epi}}\big\},
  \label{eq:negset}
\end{equation}
and use $\tau_{\mathrm{epi}}=3\,\mathrm{px}$ by default. At every layer,
their loss is
\begin{equation}
  \mathcal{L}^{lj,\ell}_{\rm neg}=
  -\frac{1}{|\mathcal{N}^{lj}|}\sum_{(p,q)\in\mathcal{N}^{lj}}
  \log(1-P^{lj,\ell}_{pq}),
  \label{eq:negative}
\end{equation}
and is zero for an empty set. Mining at the final layer focuses on mistakes
that the matcher would actually output, while supervising the same set at
every layer propagates the corrective signal throughout the matcher.

Each matcher layer is trained with the LoMa dual-softmax matching loss,
matchability loss, and the geometric negative loss. With
$\lambda_{\mathrm{neg}}=1/(N-1)$, the objective is
\begin{align}
  \mathcal{L}_{\mathrm{matcher}}&=\frac{1}{L}\sum_{\ell=1}^L
  \Big(\mathcal{L}^{\ell}_{\rm glue}+
  \mathcal{L}^{\ell}_{\rm matchability}\nonumber\\
  &\qquad\qquad+
  \lambda_{\mathrm{neg}}\sum_{j=1}^{N-1}\mathcal{L}^{lj,\ell}_{\rm neg}\Big).
  \label{eq:loss}
\end{align}
The cross-attention layers are jointly supervised over all neighbours. The
foundation encoder stays frozen; we train the descriptor head $g_\psi$ and
matcher attention layers. Section~\ref{sec:experiments} isolates the effect
of GT filtering and the negative loss.

\section{Experiments}
\label{sec:experiments}

\subsection{Protocol and Implementation}

We evaluate multi-view matching on 25-frame sets from HyperSim, ScanNet++,
Mip-NeRF 360, and Tanks and
Temples~\citep{roberts2021hypersim,yeshwanth2023scannetpp,barron2022mipnerf360,knapitsch2017tanks}.
HyperSim and ScanNet++ provide challenging indoor geometry and appearance;
Mip-NeRF 360 and Tanks and Temples add outdoor and large-scale scenes. The
encoder and view-count ablations additionally use MegaDepth-X and
TerraSky3D~\citep{li2026megadepthx,durso2026terrasky3d} to probe transfer
beyond the four primary benchmarks. The controlled descriptor-backbone
ablation also uses AerialMD (AerialMegaDepth) evaluation
pairs~\citep{vuong2025aerialmegadepth}. In that ablation, overlap $o$ is
computed following the definition of RoMa
v2~\citep{edstedt2026romav2}, and we report subsets with $o>0.35$ and
$o>0.05$.

For every 25-frame set, frame 0 is the star centre and frames 1--24 give 24
centre--neighbour edges. This protocol evaluates matching as it is used in a
local multi-view subproblem rather than sampling unrelated pairs. For a
multi-view method that accepts $K+1<25$ images, we partition the 24
neighbours into groups of size $K$. The groups cover exactly the same
centre--neighbour edges as a two-view baseline, so any difference cannot be
attributed to easier pair selection. MV-RoMa exhausts memory at 25 views and
is evaluated with 13-view groups under this same coverage rule.

For each edge, we estimate relative pose with PoseLib
RANSAC~\citep{larsson2020poselib} at a 2.5\,px threshold. We report
AUC@$1^\circ/5^\circ/10^\circ$, where the pose error combines rotation and
translation errors. The strict $1^\circ$ threshold exposes precise
correspondence failures that may be concealed by coarse thresholds. Each
baseline is evaluated at the input resolution prescribed by its original
paper; \method is reported at DA3-G encoder resolutions 504 and 784.

Unless stated otherwise, \method uses a frozen DA3-G encoder, LoMa-L
initialization, and 4096 DaD keypoints per image. The matcher architecture
and optimization hyperparameters otherwise follow LoMa-L. Unlike LoMa, we
jointly optimize the trainable descriptor components and matcher end-to-end,
while keeping the geometry foundation encoder frozen. All reported variants
are trained for 100k iterations with batch size 1. The GT/loss ablation uses
fixed four-view training. The encoder and view-count experiments use dynamic
2--8-view training, testing whether a model exposed to variable local
context can generalize to substantially larger stars.

\begin{table*}[t]
  \centering
  \caption{Relative pose AUC on 25-view star graphs; each cell reports
  AUC@$1^\circ/5^\circ/10^\circ$ in that order. MV-RoMa uses 13 views owing
  to out-of-memory at 25 views. Best and second-best values are bold and
  underlined. Average is the macro-average over the four benchmarks at each
  threshold; T\&T denotes Tanks and Temples.}
  \label{tab:sota}
  \small
  \setlength{\tabcolsep}{5pt}
  \begin{tabular}{@{}lccccc@{}}
    \toprule
    Method & HyperSim & ScanNet++ & Mip-NeRF 360 & T\&T & Average\\
    \midrule
    \multicolumn{6}{@{}l}{\emph{Two-view}}\\
    LightGlue &20.20/47.34/55.72&22.84/61.83/74.49&51.28/83.46/90.02&52.42/84.69/91.08&36.69/69.33/77.83\\
    LoMa-L &51.99/77.00/82.86&37.68/\underline{75.57}/\underline{85.14}&71.69/93.25/96.50&\textbf{68.22}/\textbf{91.96}/\textbf{95.74}&57.40/84.44/90.06\\
    RoMa &45.06/64.82/69.78&36.81/73.88/83.70&\textbf{75.42}/93.27/95.88&66.56/90.99/95.01&55.96/80.74/86.09\\
    RoMa v2 &52.93/\textbf{80.65}/\textbf{86.92}&33.02/73.33/84.58&\underline{74.81}/\textbf{94.16}/\textbf{97.00}&67.06/91.32/95.37&56.95/84.87/\underline{90.97}\\
    \midrule
    \multicolumn{6}{@{}l}{\emph{Multi-view}}\\
    MV-RoMa (13v) &37.00/58.51/64.51&35.80/71.57/81.67&74.45/92.93/95.61&66.04/90.56/94.75&53.32/78.39/84.14\\
    \method (504) &\underline{53.35}/78.99/84.76&\underline{38.03}/74.72/84.67&73.90/94.04/96.97&\underline{68.08}/\underline{91.91}/\underline{95.74}&\underline{58.34}/\underline{84.91}/90.53\\
    \method (784) &\textbf{53.90}/\underline{80.02}/\underline{85.99}&\textbf{38.44}/\textbf{75.68}/\textbf{85.33}&74.29/\underline{94.08}/\underline{96.97}&67.96/91.84/95.68&\textbf{58.65}/\textbf{85.41}/\textbf{90.99}\\
    \bottomrule
  \end{tabular}
\end{table*}

\subsection{Comparison to Matching Baselines}

Table~\ref{tab:sota} compares sparse and dense baselines under identical
centre--neighbour edge coverage. MV-RoMa is evaluated with 13 views because
its 25-view configuration exhausts available memory.

At 784 resolution, \method improves AUC@$1^\circ$ over LoMa-L by 1.91 points
on HyperSim and 0.76 points on ScanNet++. The gains are consistent with the
value of jointly contextualizing centre features and using cross-view
foundation features before assignment, although this comparison does not
separate the two factors. It also achieves the best macro-average at all
three thresholds (58.65/85.41/90.99), directly supporting the aggregate
claim in the abstract. The 784 model obtains a higher strict HyperSim score
than the 504 model (53.90 versus 53.35); on the remaining benchmarks, the
two resolutions are close.

The comparison also identifies meaningful boundaries. On Mip-NeRF 360, dense
two-view RoMa remains strongest at AUC@$1^\circ$, indicating that fine dense
localization can still dominate for this scene distribution. At the coarser
5- and 10-degree thresholds, \method is the strongest multi-view method. On
Tanks and Temples, LoMa-L retains the best score and \method stays within a
small margin. Rather than suggesting that multi-view reasoning universally
replaces pairwise dense matching, these results support it as a
complementary sparse design for star-structured geometric estimation.

\subsection{Ablations}

\paragraph{Ground-truth filtering and negative supervision.}
Table~\ref{tab:gtneg} shows that both components improve matching. The
geometric negative loss produces the larger HyperSim AUC@$1^\circ$ change
(52.84 to 51.04 when removed), compared with 52.84 to 51.93 without GT
filtering. The two modifications address different failures: filtering
prevents invalid dense warps from becoming positives, whereas the negative
term targets the high-confidence false matches that remain after assignment.
Their improvements on both HyperSim and Mip-NeRF 360 show that the effect is
observed beyond the synthetic indoor setting. The larger strict-threshold
drop without the negative term is consistent with epipolar-inconsistent
outliers degrading robust pose estimation.

\begin{table}[t]
  \centering
  \caption{GT filtering and geometric negative loss (DA3-G; four-view
  training; test at 784). Best and second-best values are bold and
  underlined.}
  \label{tab:gtneg}
  \small
  \setlength{\tabcolsep}{3.4pt}
  \begin{tabular}{@{}lcccccc@{}}
    \toprule
    &\multicolumn{3}{c}{HyperSim}&\multicolumn{3}{c}{Mip-NeRF 360}\\
    \cmidrule(lr){2-4}\cmidrule(lr){5-7}
    Variant&@1&@5&@10&@1&@5&@10\\
    \midrule
    Full&\textbf{52.84}&\textbf{78.42}&\textbf{84.34}&\textbf{73.85}&\textbf{93.96}&\textbf{96.92}\\
    w/o GT filtering&\underline{51.93}&\underline{77.75}&\underline{83.77}&\underline{73.69}&\underline{93.85}&\underline{96.83}\\
    w/o negative loss&51.04&76.96&83.19&72.72&93.42&96.55\\
    \bottomrule
  \end{tabular}
\end{table}

\paragraph{Descriptor backbone.}
Table~\ref{tab:backbone} isolates the frozen descriptor input while keeping
the training data, resolution, star matcher, losses, and optimization
configuration fixed. DA3-G consistently outperforms both DINOv2 variants.
Compared with DINOv2-G, it improves AUC@$1^\circ$ by 1.16, 11.44, 0.41, and
0.65 points across the four test subsets, respectively. The smaller
differences between DINOv2-G and DINOv2-L indicate that scaling the frozen
2D backbone alone does not explain the improvement.

\begin{table}[t]
  \centering
  \caption{Controlled frozen-backbone ablation at resolution 784. We report
  AUC@$1^\circ$ at two RoMa v2 overlap
  thresholds~\citep{edstedt2026romav2}. Best and second-best values are bold
  and underlined.}
  \label{tab:backbone}
  \small
  \setlength{\tabcolsep}{5pt}
  \begin{tabular}{@{}lcccc@{}}
    \toprule
    &\multicolumn{2}{c}{HyperSim}&\multicolumn{2}{c}{AerialMD}\\
    \cmidrule(lr){2-3}\cmidrule(lr){4-5}
    Backbone&$o>0.35$&$o>0.05$&$o>0.35$&$o>0.05$\\
    \midrule
    DA3-G&\textbf{69.27}&\textbf{61.14}&\textbf{77.97}&\textbf{48.16}\\
    DINOv2-G&\underline{68.11}&\underline{49.70}&\underline{77.56}&\underline{47.51}\\
    DINOv2-L&67.72&48.29&77.33&47.08\\
    \bottomrule
  \end{tabular}
\end{table}

\paragraph{Foundation encoder.}
Table~\ref{tab:encoder} evaluates frozen geometry encoders under the same
2--8-view training setup. DA3-G is best on all reported metrics, while
VGGT-$\Omega$ is close on MegaDepth-X and TerraSky3D. Thus, the benefits are
not solely from the multi-view matcher: the quality of the frozen
geometry-aware token representation matters. In particular, DA3-G exceeds
the other encoders on HyperSim AUC@$1^\circ$ by 1.62--2.40 points. This
indicates that the choice of frozen representation remains consequential
after the same trainable descriptor head and matcher are applied. The result
supports the use of a geometry-aware encoder as an input to a lightweight
local matching head, without establishing that any individual property of
DA3-G is solely responsible for the difference.

\begin{table*}[t]
  \centering
  \caption{Frozen foundation encoder ablation for \method (2--8 train views,
  25 test views, resolution 784). Best and second-best values are bold and
  underlined.}
  \label{tab:encoder}
  \small
  \setlength{\tabcolsep}{6pt}
  \begin{tabular}{@{}lccccccccc@{}}
    \toprule
    &\multicolumn{3}{c}{HyperSim}&\multicolumn{3}{c}{MegaDepth-X}&\multicolumn{3}{c}{TerraSky3D}\\
    \cmidrule(lr){2-4}\cmidrule(lr){5-7}\cmidrule(lr){8-10}
    Encoder&@1&@5&@10&@1&@5&@10&@1&@5&@10\\
    \midrule
    DA3-G&\textbf{53.90}&\textbf{80.02}&\textbf{85.99}&\textbf{18.88}&\textbf{52.97}&\textbf{65.77}&\textbf{55.53}&\textbf{83.88}&\textbf{90.04}\\
    $\pi^3$&51.50&76.56&82.57&18.37&52.78&65.42&53.77&82.81&89.45\\
    VGGT&\underline{52.28}&\underline{78.17}&\underline{84.26}&18.05&52.22&64.70&53.21&81.92&88.56\\
    VGGT-$\Omega$&52.07&77.70&83.46&\underline{18.43}&\underline{52.91}&\underline{65.58}&\underline{55.21}&\underline{83.38}&\underline{89.78}\\
    \bottomrule
  \end{tabular}
\end{table*}

\begin{figure}[t]
  \centering
  \includegraphics[width=\linewidth]{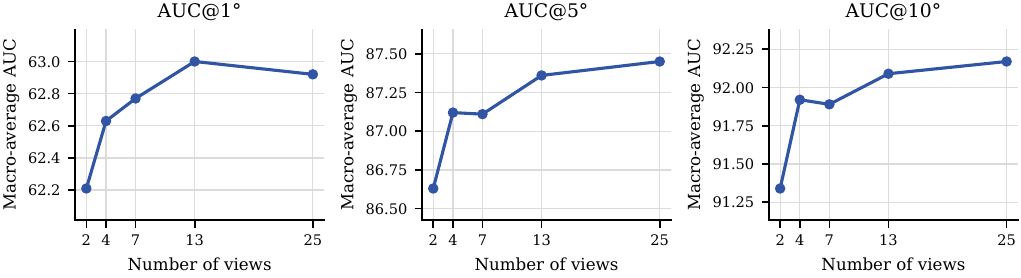}
  \caption{Macro-average relative pose AUC from Table~\ref{tab:views} versus
  the number of inference views. Each panel averages the corresponding AUC
  threshold over HyperSim, Mip-NeRF 360, Tanks and Temples, and TerraSky3D.}
  \label{fig:view-count-auc}
\end{figure}

\paragraph{Number of inference views.}
Table~\ref{tab:views} varies the input-star size at test time and reports
the dataset-level scores, while Figure~\ref{fig:view-count-auc} isolates
their macro-average trend. Under the fixed-edge protocol, the view count
changes only how many neighbours are processed jointly in each star. We
observe a clear and consistent performance drop at two views: the
macro-average AUC falls to 62.21/86.63/91.34 at $1^\circ/5^\circ/10^\circ$,
approximately 0.8 points below the best multi-view result at every
threshold. In the two-view setting, each star contains only one centre and
one neighbour, so star matching degenerates to pair matching and can no
longer aggregate evidence across multiple neighbours. Since the number of
evaluated pairs remains unchanged, the observed drop isolates the loss of
joint multi-view interaction rather than a change in evaluation scale,
providing direct empirical evidence for the benefit of star-graph matching
over its pairwise counterpart.

\begin{table*}[t]
  \centering
  \caption{Inference view count for DA3-G \method trained with 2--8 views at
  resolution 784; each cell reports AUC@$1^\circ/5^\circ/10^\circ$. Average
  is the macro-average over the four benchmarks at each threshold. Best and
  second-best values are bold and underlined; T\&T denotes Tanks and
  Temples.}
  \label{tab:views}
  \small
  \setlength{\tabcolsep}{5pt}
  \begin{tabular}{@{}lccccc@{}}
    \toprule
    \# views & HyperSim & Mip-NeRF 360 & T\&T & TerraSky3D & Average\\
    \midrule
    2&52.51/77.63/83.29&73.96/94.02/96.95&67.63/\underline{91.86}/\underline{95.73}&54.76/83.01/89.37&62.21/86.63/91.34\\
    4&53.79/79.43/85.68&\underline{74.31}/94.02/96.95&67.85/\textbf{91.89}/\textbf{95.78}&54.58/83.13/89.29&62.63/87.12/91.92\\
    7&53.90/79.13/85.14&\textbf{74.32}/\underline{94.06}/\underline{96.97}&67.89/91.82/95.70&54.97/83.42/89.75&62.77/87.11/91.89\\
    13&\textbf{54.47}/\textbf{80.12}/\underline{85.88}&74.20/94.04/96.96&\underline{67.92}/91.82/95.66&\underline{55.39}/\underline{83.47}/\underline{89.85}&\textbf{63.00}/\underline{87.36}/\underline{92.09}\\
    25&\underline{53.90}/\underline{80.02}/\textbf{85.99}&74.29/\textbf{94.08}/\textbf{96.97}&\textbf{67.96}/91.84/95.68&\textbf{55.53}/\textbf{83.88}/\textbf{90.04}&\underline{62.92}/\textbf{87.45}/\textbf{92.17}\\
    \bottomrule
  \end{tabular}
\end{table*}

Beyond two views, the gains depend on the dataset and threshold. HyperSim
improves from 52.51 at two views to 54.47 at 13 views, while TerraSky3D
improves through 25 views. The trend is not monotonic: the HyperSim
$1^\circ$ score is lower at 25 than at 13 views despite improving at the
coarser thresholds. Additional observations may provide complementary
support but can also introduce distractors or harder baselines. The results
therefore demonstrate extrapolation beyond the 2--8-view training range
without implying that every added view is beneficial.

\paragraph{Matching complexity.}
Let a star contain $V$ views, each with $K$ keypoints, and let the matcher
hidden dimension be $d$. Cross-attention has the same quadratic cost in both
implementations because $(V-1)(K\!\times\!K)=K\!\times\!((V-1)K)$. The
savings instead come from avoiding repeated centre-side self-attention,
linear projections, and feed-forward networks. Ignoring LayerNorm, RoPE, and
softmax, one LoMa-L pair costs approximately $76Kd^2+16K^2d$ FLOPs per
layer. Across $L$ layers, sequential pairwise and star matching require
\begin{align}
  C_{\rm pair} &\approx L(V-1)(76Kd^2+16K^2d),\\
  C_{\rm star} &\approx L\big(38VKd^2+(12V-8)K^2d\big).
  \label{eq:complexity}
\end{align}
With $K=2048$, $d=512$, and $L=9$, their theoretical compute ratio is
$C_{\rm pair}/C_{\rm star}=70(V-1)/(43V-16)$. Table~\ref{tab:runtime}
compares this estimate with matcher-only wall-clock time.

\begin{table}[t]
  \centering
  \caption{Transformer compute and matcher-only runtime for nine LoMa-L
  layers. FLOPs count a multiply--add as two operations. Runtime uses one
  NVIDIA A800, batch size 1, and 2048 keypoints; values average 46 of 50
  trials after dropping the two fastest and two slowest.}
  \label{tab:runtime}
  \small
  \setlength{\tabcolsep}{3.2pt}
  \begin{tabular}{@{}rrcccccc@{}}
    \toprule
    & & \multicolumn{3}{c}{Theory (TFLOPs)} & \multicolumn{3}{c}{A800 (ms)}\\
    \cmidrule(lr){3-5}\cmidrule(lr){6-8}
    Views & Edges & Pair & Star & Sp.\ & Pair & Star & Sp.\\
    \midrule
    2  & 1  & 0.676 & 0.676 & $1.00\times$ & 20.95  & 21.18  & $0.99\times$\\
    4  & 3  & 2.029 & 1.508 & $1.35\times$ & 57.98  & 37.15  & $1.56\times$\\
    7  & 6  & 4.059 & 2.754 & $1.47\times$ & 113.81 & 60.50  & $1.88\times$\\
    13 & 12 & 8.117 & 5.247 & $1.55\times$ & 237.60 & 105.78 & $2.25\times$\\
    25 & 24 & 16.235 & 10.234 & $1.59\times$ & 446.96 & 203.09 & $2.20\times$\\
    \bottomrule
  \end{tabular}
\end{table}

At 25 views, star matching reduces the Transformer FLOP count by $37.0\%$,
an architectural saving that is independent of whether the pairwise edges are
executed sequentially or batched: batching can amortize launch overhead, but
it still repeats the centre-side self-attention, projections, and
feed-forward networks for every edge. Relative to sequential pairwise
evaluation, the measured latency is reduced by $54.6\%$ ($2.20\times$
speedup). This measured gain is larger than the theoretical $1.59\times$
compute ratio because the 24 pairwise calls additionally incur repeated
Python/module and kernel-launch overhead, whereas the star implementation
uses larger contiguous matrices with better GPU utilization. The two-view
outputs are identical, providing an implementation sanity check. These
compute and latency savings come at the cost of higher peak memory: at 25
views, matcher-only peak memory is 979\,MiB for the star matcher versus
176\,MiB for sequential pairwise evaluation.

\section{Reproducibility and Limitations}

We describe the architecture, loss, evaluation topology, RANSAC setting,
keypoint budget, frozen/trainable components, and reported training
schedules in the main paper, with further detail in
Appendix~\ref{app:implementation}. The method builds directly on publicly
available LoMa-style matching components and multi-view 3D foundation
encoders. We will release the implementation and evaluation configuration
upon publication.

\method depends on the geometric quality of the frozen foundation features;
textureless or uncertain regions can still yield sparse or ambiguous
matches. The centre--neighbour topology does not directly exchange
information between neighbours, and the memory requirement still grows with
the number of keypoints and views. Our 2--8-view training range is also
smaller than the 25-view evaluation star. Although the view-count ablation
shows useful extrapolation, it is not uniformly monotonic and does not
establish that all additional observations are beneficial. Learning view
selection or confidence-aware neighbour aggregation is a promising direction
for larger stars. Finally, we evaluate the matcher through relative pose
estimation; its interaction with complete global SfM systems is outside the
scope of this work.

\section{Conclusion}

We presented \method to resolve a mismatch in current geometric pipelines:
pairwise assignment discards evidence and foundation features that are
already available across a view collection. \method reuses frozen
geometry-aware tokens and shares the centre representation across a star,
while preserving ordinary sparse pairwise outputs for robust pose
estimation. Across four benchmarks, the 784 model attains the strongest
macro-average AUC among the compared methods at all three thresholds, and
its 25-view matcher is $2.20\times$ faster than sequential pairwise
evaluation. The ablations further show that the gains depend on both the
frozen representation and the proposed supervision, while the non-monotonic
view-count results clarify that additional context is useful but not
uniformly beneficial.

\clearpage
\appendix
\onecolumn
\raggedbottom

\section{Implementation Details}
\label{app:implementation}

At inference, we use 4096 DaD keypoints per image. The trainable VGG stream,
convolutional refiner, descriptor projection, and matcher are optimized
jointly, while the foundation encoder remains frozen. Geometric negatives
use a 3\,px epipolar threshold and weight $1/(N-1)$. The fixed-view GT/loss
ablation uses LoMa-L initialization, four views, batch size one, and 100k
iterations at resolution 784. Encoder and view-count ablations sample 2--8
views; evaluation covers all 24 centre--neighbour edges in each 25-frame
set.

\paragraph{Training datasets.}
Table~\ref{tab:app-datasets} lists the final DA3-G data mixture. Intervals
denote visible-overlap ranges. In addition to datasets cited in the main
text, the mixture uses MegaDepth, BlendedMVS, and
TartanAir~\citep{li2018megadepth,yao2020blendedmvs,wang2020tartanair};
MegaDepth (Re-MVS) follows the LoMa
recomputation~\citep{nordstrom2026loma}.

\begin{table}[H]
  \centering
  \caption{Training data mixture for the final DA3-G model. MVS denotes
  multi-view stereo; Sample w.\ and Neg.\ w.\ are dataset-sampling and
  geometric negative-loss weights.}
  \label{tab:app-datasets}
  \small
  \setlength{\tabcolsep}{6pt}
  \begin{tabular}{@{}lllclc@{}}
    \toprule
    Dataset & Domain & GT source & Sample w. & Overlap sampling & Neg. w. \\
    \midrule
    MegaDepth & Outdoor & MVS & 1.0 & 5--35 / 35--100 (0.5 each) & 1.0 \\
    MegaDepth (Re-MVS) & Outdoor & MVS & 1.0 & 1--35 / 35--100 (0.5 each) & 1.0 \\
    MegaDepth-X & Outdoor & MVS & 1.0 & 1--35 / 35--100 (0.5 each) & 0.01 \\
    HyperSim & Indoor & Graphics & 1.0 & 5--50 / 50--100 (0.5 each) & 0.01 \\
    ScanNet++ & Indoor & Mesh & 0.5 & 20--100 only & 0.1 \\
    BlendedMVS & Aerial & Mesh & 1.0 & 5--20 / 20--100 (0.5 each) & 1.0 \\
    AerialMD & Aerial & MVS & 1.0 & 5--20 / 20--100 (0.5 each) & 1.0 \\
    TerraSky3D & Aerial & MVS & 0.5 & Open-source JSON split & 0.01 \\
    TartanAir & Outdoor & Graphics & 0.5 & 5--20 / 20--100 (0.25 each) & 1.0 \\
    \bottomrule
  \end{tabular}
\end{table}

\paragraph{Optimization and initialization.}
All final experiments use $784\times784$ inputs and the publicly released
LoMa-L checkpoint as initialization. The descriptor and matcher learning
rates are $10^{-4}$ and $2\times10^{-4}$, respectively. We use AdamW with
weight decay $5\times10^{-5}$ and an exponential-moving-average decay of
$0.999$. Training uses one NVIDIA A800 GPU, batch size one, and 200k
iterations: the first 100k iterations use fixed four-view stars, followed by
100k iterations with the number of views sampled uniformly from 2--8.
MegaDepth-X has sparse depth supervision, so its auxiliary matchability loss
is disabled; its geometric negative-loss weight remains as reported in
Table~\ref{tab:app-datasets}.

\clearpage
\section{Qualitative Visualizations}
\label{app:visualizations}

This appendix collects qualitative \method visualizations from the
development experiments. For each star, the original RGB views are shown
above the corresponding star-matching visualization. Examples are grouped by
dataset.
\vspace{0.3em}

\newlength{\viswidth}\setlength{\viswidth}{0.53\textwidth}
\newlength{\visright}\setlength{\visright}{0.13\textwidth}
\par\addvspace{0.6em}\noindent
\begin{minipage}{\textwidth}
{\small\sffamily\bfseries Blended MVS}\par\vspace{0.1em}
\makebox[\linewidth][c]{%
    \begin{minipage}[c]{0.12\textwidth}\centering\textbf{\scriptsize RGB input}\end{minipage}%
    \hspace{0.01\textwidth}%
    \begin{minipage}[c]{\viswidth}\includegraphics[width=\linewidth]{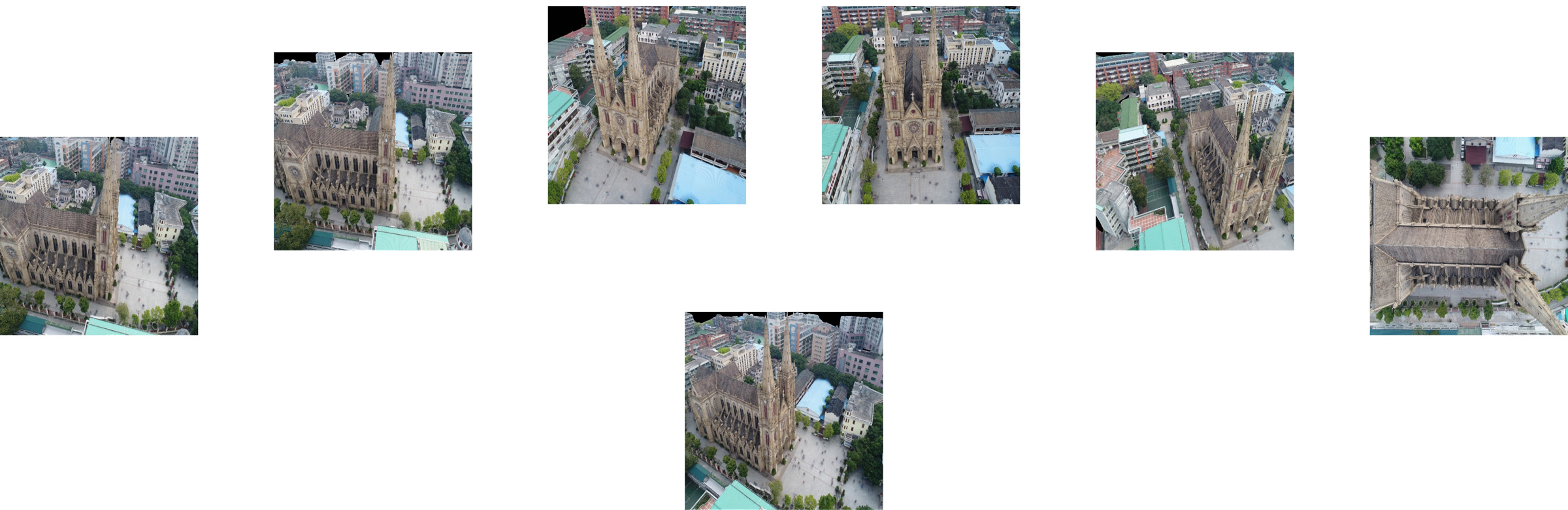}\end{minipage}%
    \hspace{\visright}}\par\vspace{-0.2em}
\makebox[\linewidth][c]{%
    \begin{minipage}[c]{0.12\textwidth}\centering\textbf{\scriptsize Star matching}\end{minipage}%
    \hspace{0.01\textwidth}%
    \begin{minipage}[c]{\viswidth}\includegraphics[width=\linewidth]{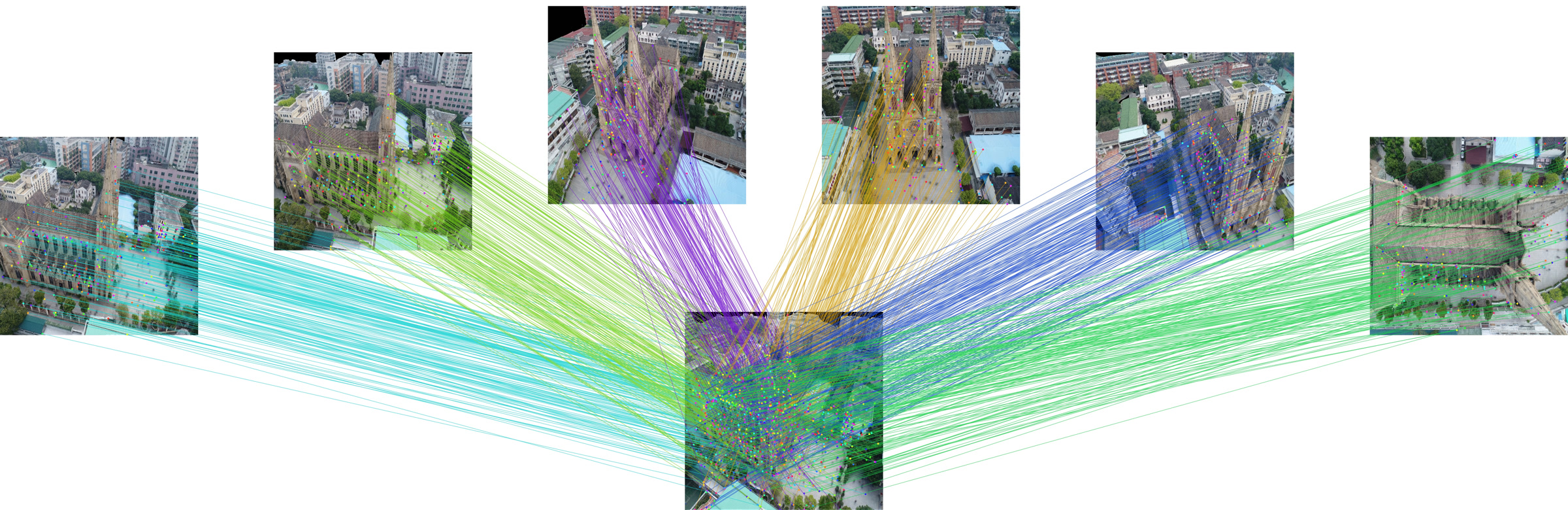}\end{minipage}%
    \hspace{\visright}}
\end{minipage}\par
\vfill
\par\addvspace{0.6em}\noindent
\begin{minipage}{\textwidth}
{\small\sffamily\bfseries HyperSim}\par\vspace{0.1em}
\makebox[\linewidth][c]{%
    \begin{minipage}[c]{0.12\textwidth}\centering\textbf{\scriptsize RGB input}\end{minipage}%
    \hspace{0.01\textwidth}%
    \begin{minipage}[c]{\viswidth}\includegraphics[width=\linewidth]{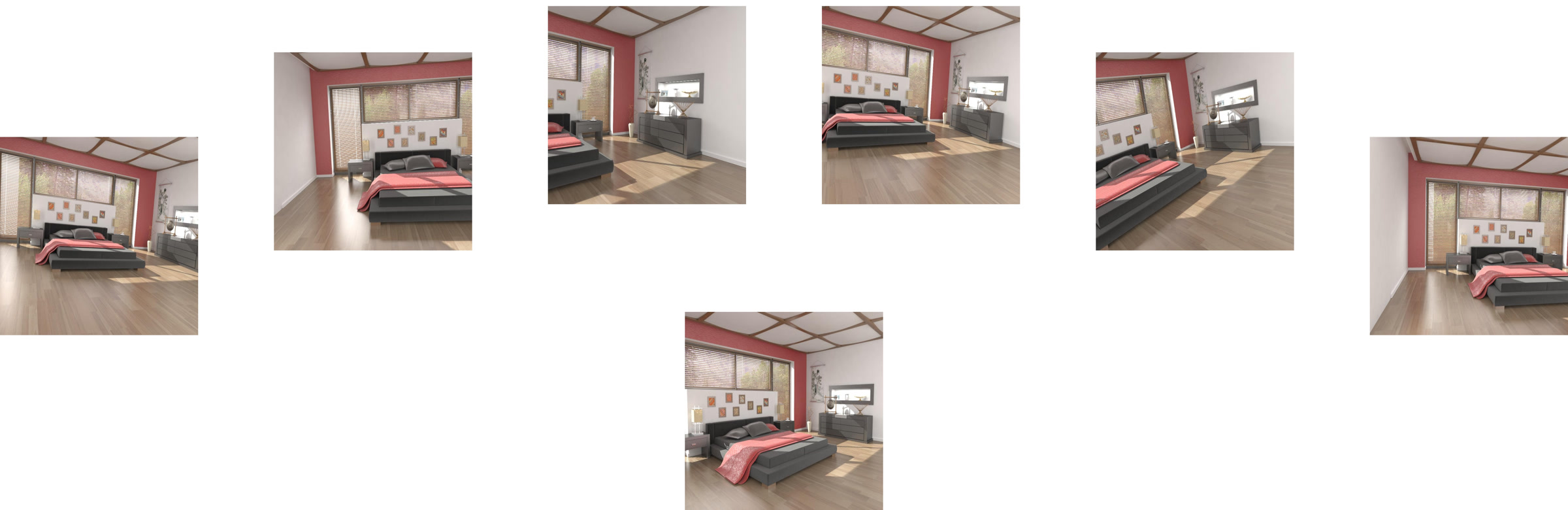}\end{minipage}%
    \hspace{\visright}}\par\vspace{-0.2em}
\makebox[\linewidth][c]{%
    \begin{minipage}[c]{0.12\textwidth}\centering\textbf{\scriptsize Star matching}\end{minipage}%
    \hspace{0.01\textwidth}%
    \begin{minipage}[c]{\viswidth}\includegraphics[width=\linewidth]{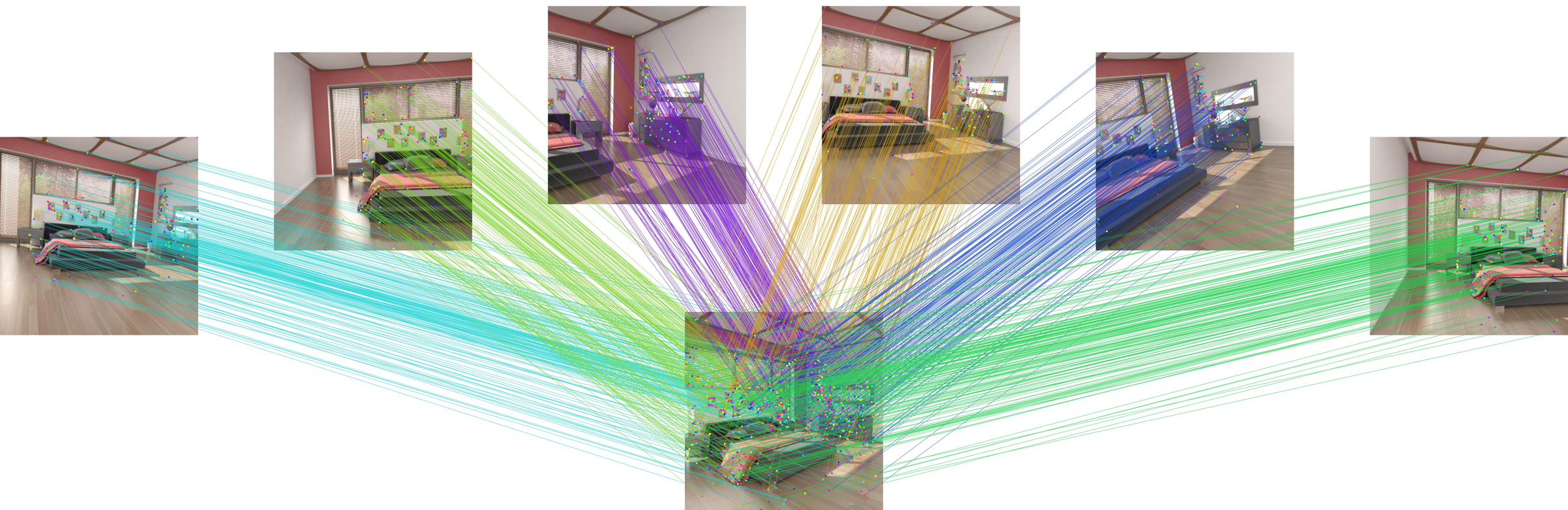}\end{minipage}%
    \hspace{\visright}}
\end{minipage}\par
\vfill
\par\addvspace{0.6em}\noindent
\begin{minipage}{\textwidth}
\makebox[\linewidth][c]{%
    \begin{minipage}[c]{0.12\textwidth}\centering\textbf{\scriptsize RGB input}\end{minipage}%
    \hspace{0.01\textwidth}%
    \begin{minipage}[c]{\viswidth}\includegraphics[width=\linewidth]{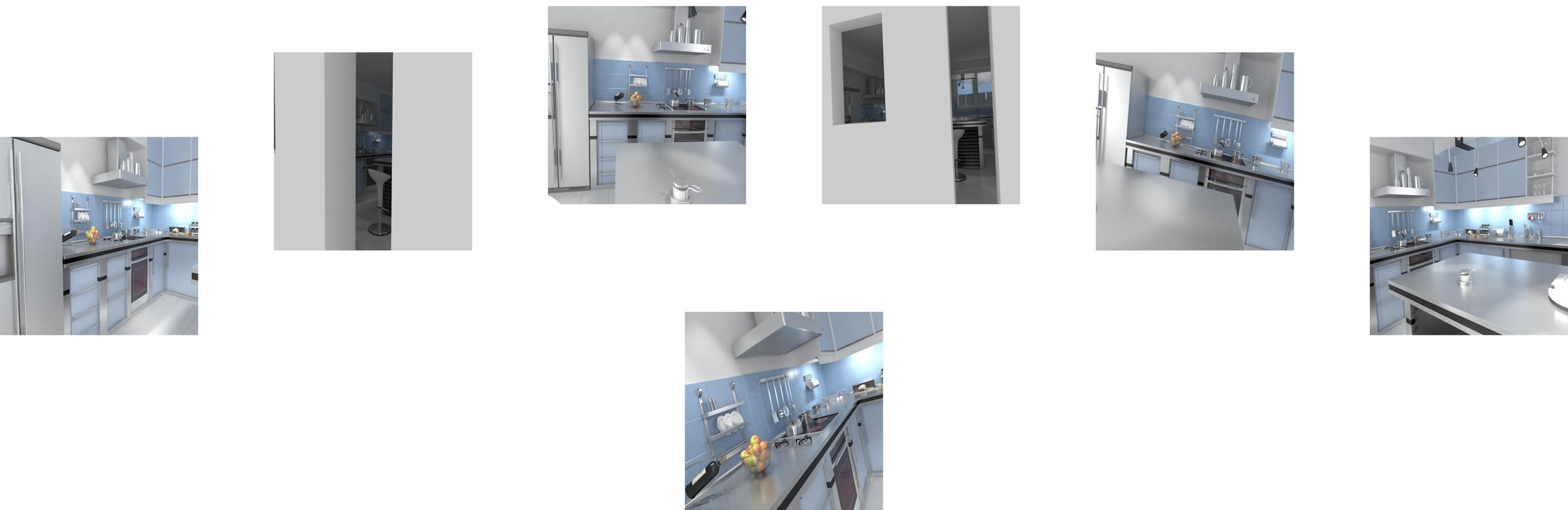}\end{minipage}%
    \hspace{\visright}}\par\vspace{-0.2em}
\makebox[\linewidth][c]{%
    \begin{minipage}[c]{0.12\textwidth}\centering\textbf{\scriptsize Star matching}\end{minipage}%
    \hspace{0.01\textwidth}%
    \begin{minipage}[c]{\viswidth}\includegraphics[width=\linewidth]{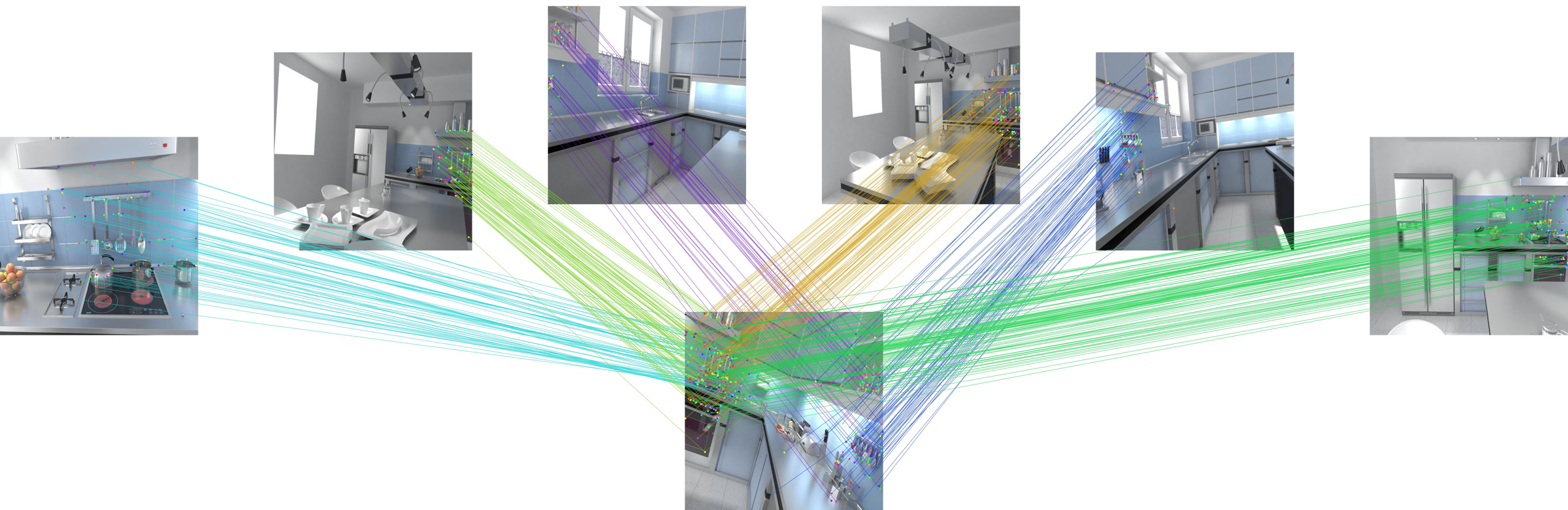}\end{minipage}%
    \hspace{\visright}}
\end{minipage}\par
\clearpage
\par\addvspace{0.6em}\noindent
\begin{minipage}{\textwidth}
{\small\sffamily\bfseries MegaDepth}\par\vspace{0.1em}
\makebox[\linewidth][c]{%
    \begin{minipage}[c]{0.12\textwidth}\centering\textbf{\scriptsize RGB input}\end{minipage}%
    \hspace{0.01\textwidth}%
    \begin{minipage}[c]{\viswidth}\includegraphics[width=\linewidth]{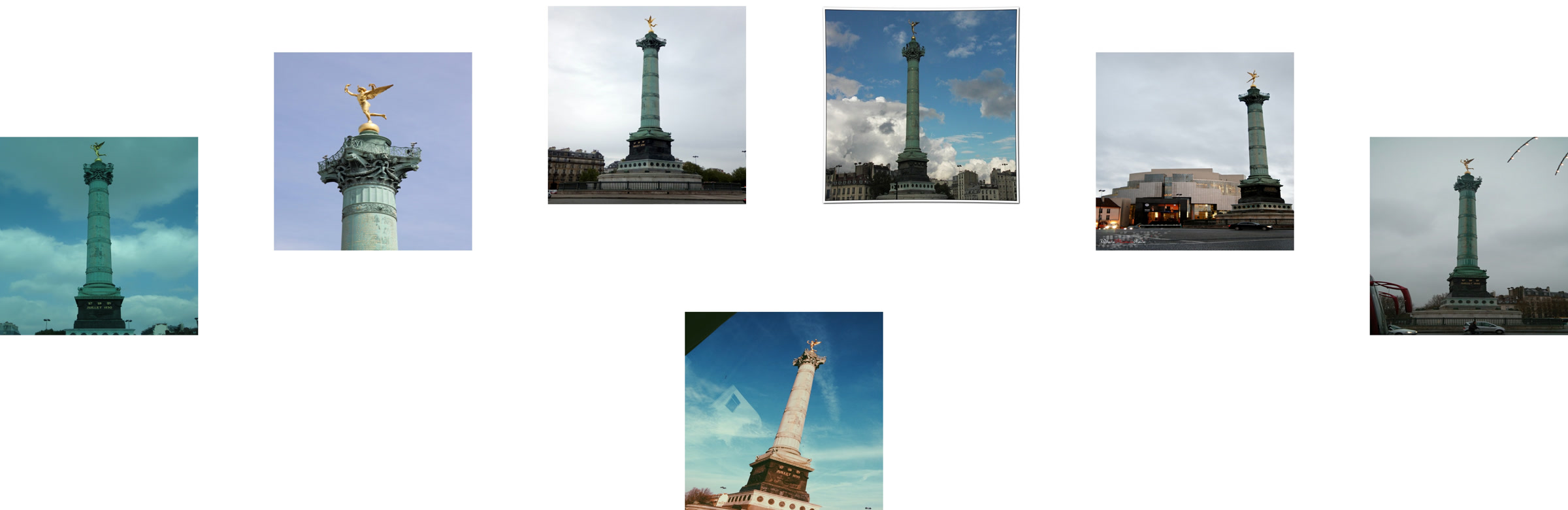}\end{minipage}%
    \hspace{\visright}}\par\vspace{-0.2em}
\makebox[\linewidth][c]{%
    \begin{minipage}[c]{0.12\textwidth}\centering\textbf{\scriptsize Star matching}\end{minipage}%
    \hspace{0.01\textwidth}%
    \begin{minipage}[c]{\viswidth}\includegraphics[width=\linewidth]{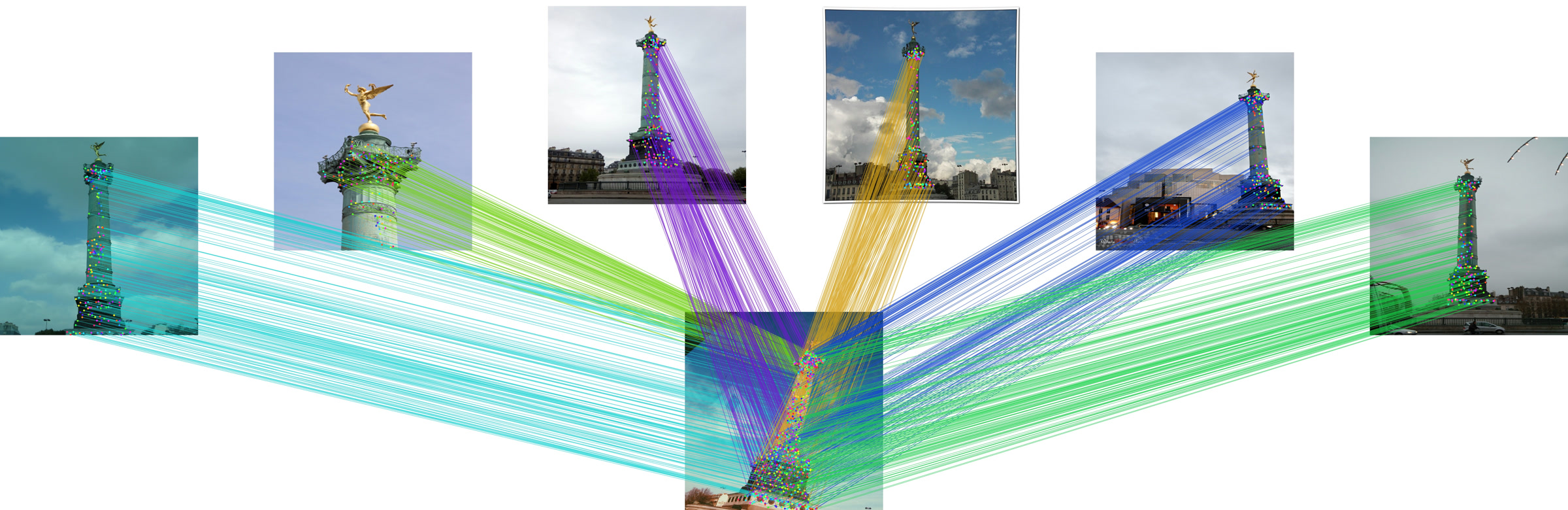}\end{minipage}%
    \hspace{\visright}}
\end{minipage}\par
\vfill
\par\addvspace{0.6em}\noindent
\begin{minipage}{\textwidth}
\makebox[\linewidth][c]{%
    \begin{minipage}[c]{0.12\textwidth}\centering\textbf{\scriptsize RGB input}\end{minipage}%
    \hspace{0.01\textwidth}%
    \begin{minipage}[c]{\viswidth}\includegraphics[width=\linewidth]{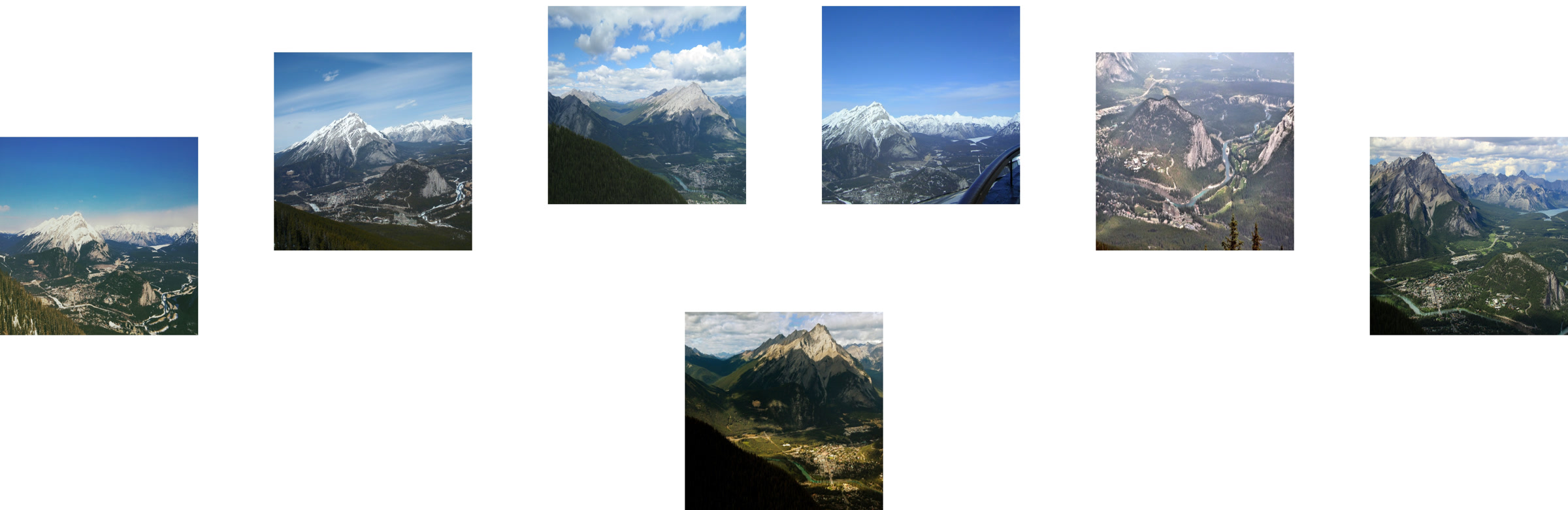}\end{minipage}%
    \hspace{\visright}}\par\vspace{-0.2em}
\makebox[\linewidth][c]{%
    \begin{minipage}[c]{0.12\textwidth}\centering\textbf{\scriptsize Star matching}\end{minipage}%
    \hspace{0.01\textwidth}%
    \begin{minipage}[c]{\viswidth}\includegraphics[width=\linewidth]{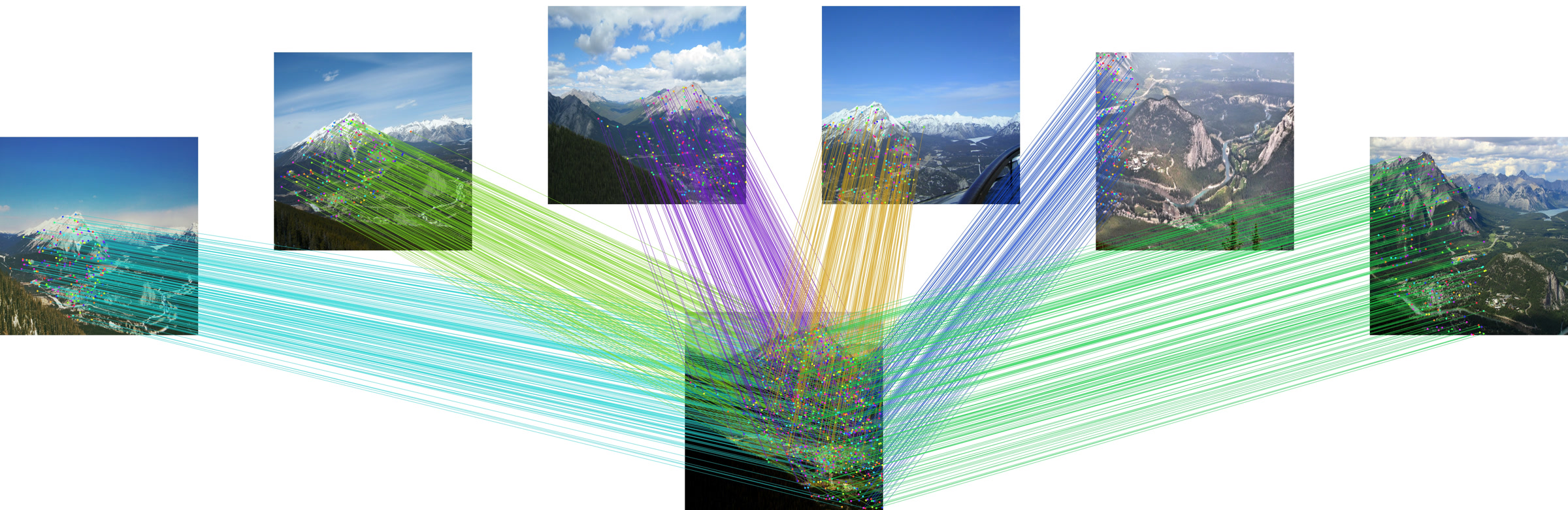}\end{minipage}%
    \hspace{\visright}}
\end{minipage}\par
\vfill
\par\addvspace{0.6em}\noindent
\begin{minipage}{\textwidth}
{\small\sffamily\bfseries MegaDepth-X}\par\vspace{0.1em}
\makebox[\linewidth][c]{%
    \begin{minipage}[c]{0.12\textwidth}\centering\textbf{\scriptsize RGB input}\end{minipage}%
    \hspace{0.01\textwidth}%
    \begin{minipage}[c]{\viswidth}\includegraphics[width=\linewidth]{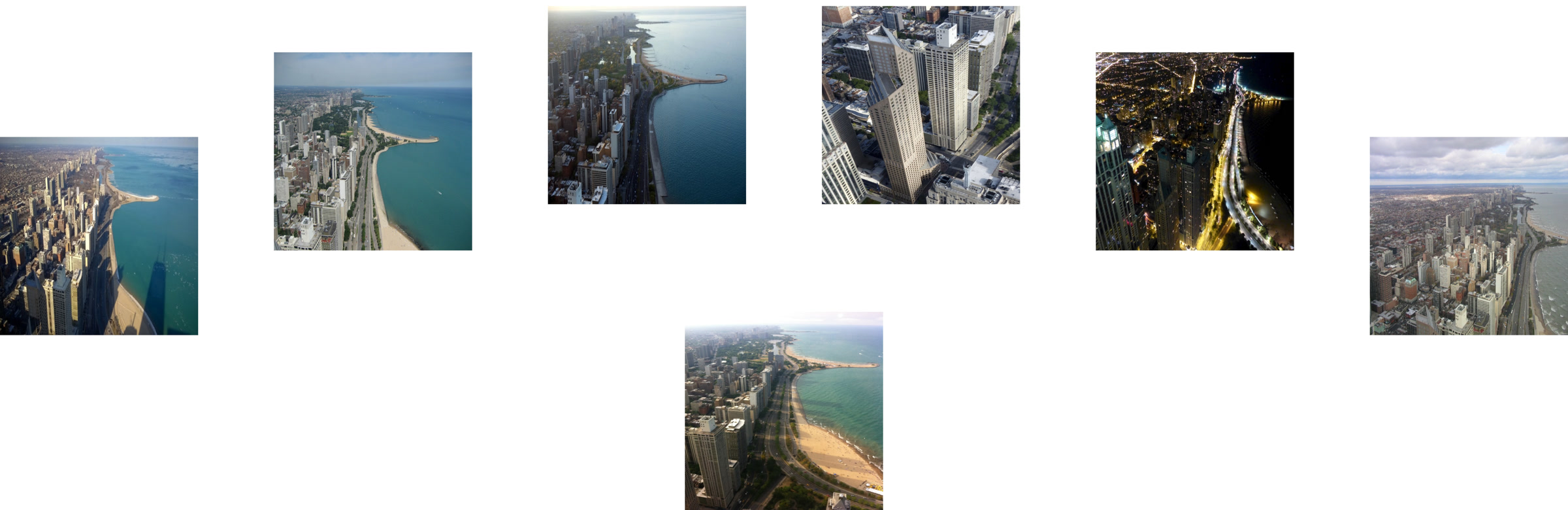}\end{minipage}%
    \hspace{\visright}}\par\vspace{-0.2em}
\makebox[\linewidth][c]{%
    \begin{minipage}[c]{0.12\textwidth}\centering\textbf{\scriptsize Star matching}\end{minipage}%
    \hspace{0.01\textwidth}%
    \begin{minipage}[c]{\viswidth}\includegraphics[width=\linewidth]{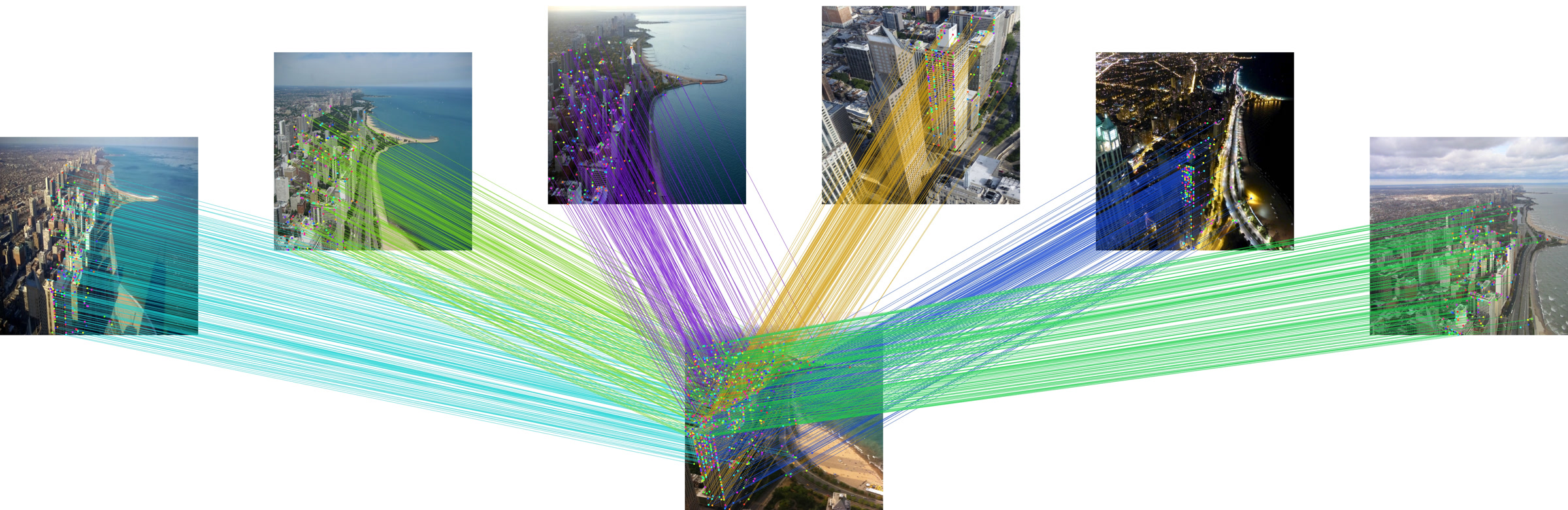}\end{minipage}%
    \hspace{\visright}}
\end{minipage}\par
\clearpage
\par\addvspace{0.6em}\noindent
\begin{minipage}{\textwidth}
{\small\sffamily\bfseries TerraSky3D}\par\vspace{0.1em}
\makebox[\linewidth][c]{%
    \begin{minipage}[c]{0.12\textwidth}\centering\textbf{\scriptsize RGB input}\end{minipage}%
    \hspace{0.01\textwidth}%
    \begin{minipage}[c]{\viswidth}\includegraphics[width=\linewidth]{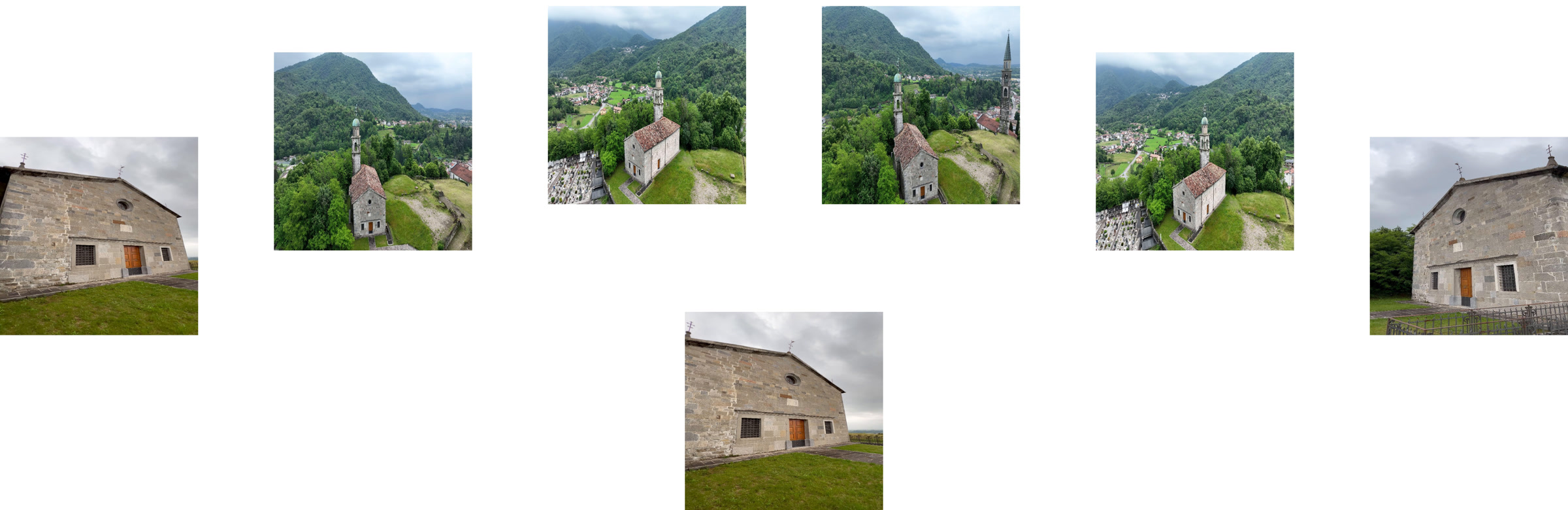}\end{minipage}%
    \hspace{\visright}}\par\vspace{-0.2em}
\makebox[\linewidth][c]{%
    \begin{minipage}[c]{0.12\textwidth}\centering\textbf{\scriptsize Star matching}\end{minipage}%
    \hspace{0.01\textwidth}%
    \begin{minipage}[c]{\viswidth}\includegraphics[width=\linewidth]{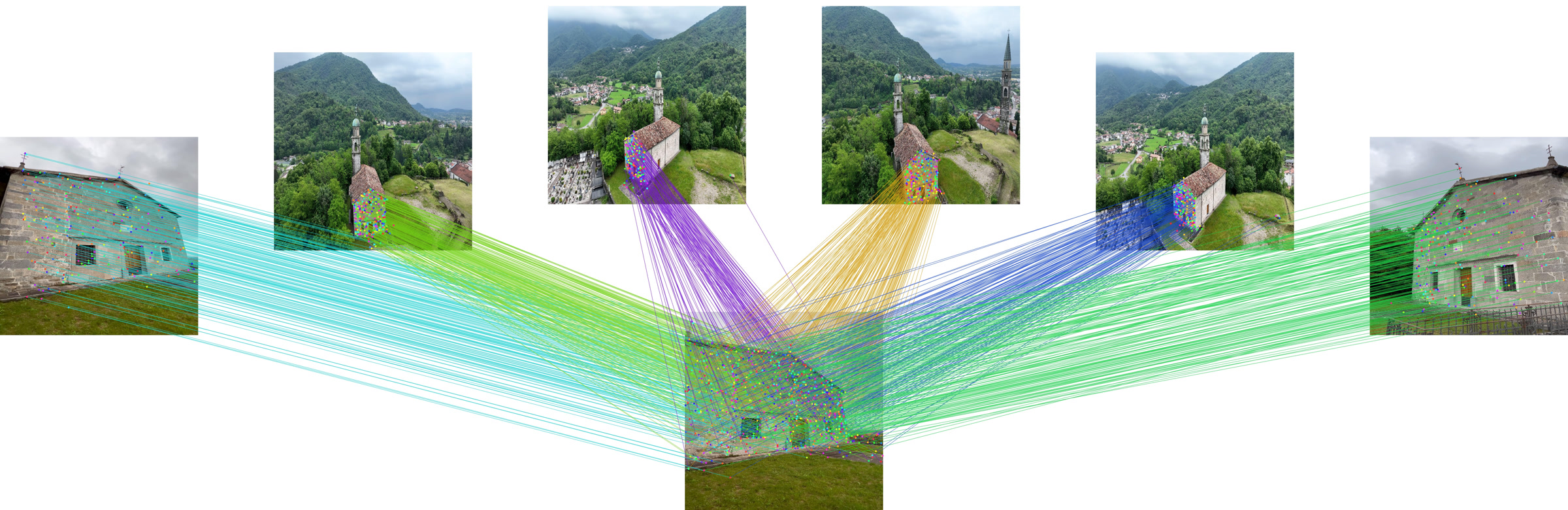}\end{minipage}%
    \hspace{\visright}}
\end{minipage}\par
\vfill
\par\addvspace{0.6em}\noindent
\begin{minipage}{\textwidth}
\makebox[\linewidth][c]{%
    \begin{minipage}[c]{0.12\textwidth}\centering\textbf{\scriptsize RGB input}\end{minipage}%
    \hspace{0.01\textwidth}%
    \begin{minipage}[c]{\viswidth}\includegraphics[width=\linewidth]{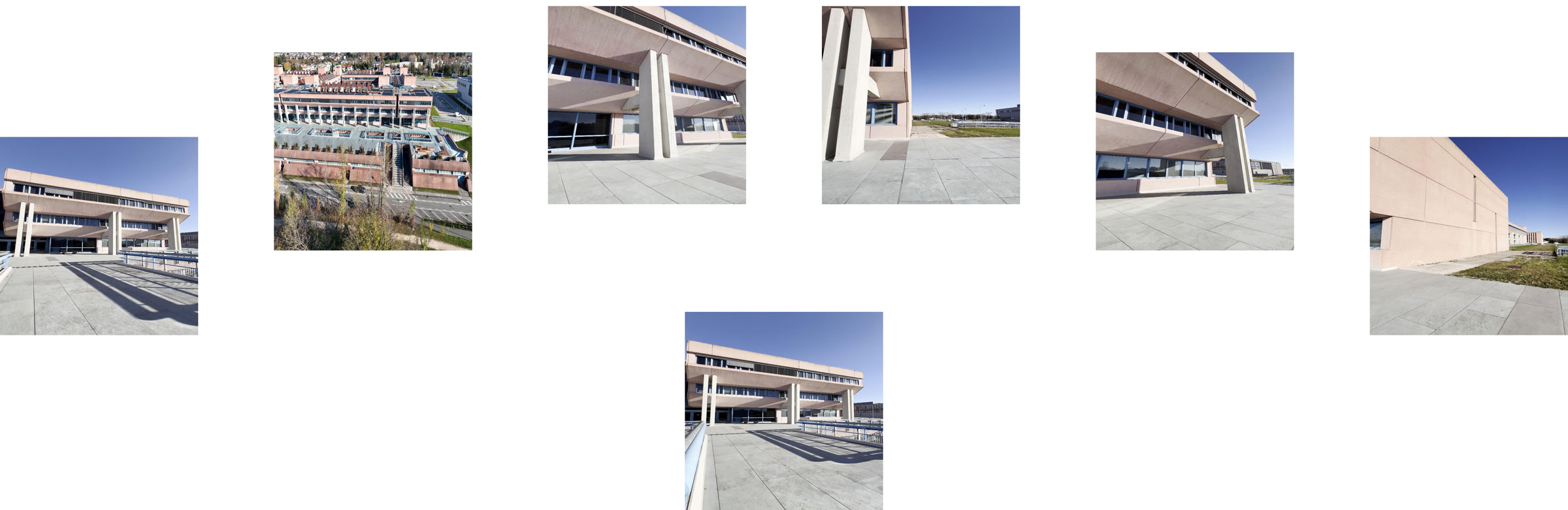}\end{minipage}%
    \hspace{\visright}}\par\vspace{-0.2em}
\makebox[\linewidth][c]{%
    \begin{minipage}[c]{0.12\textwidth}\centering\textbf{\scriptsize Star matching}\end{minipage}%
    \hspace{0.01\textwidth}%
    \begin{minipage}[c]{\viswidth}\includegraphics[width=\linewidth]{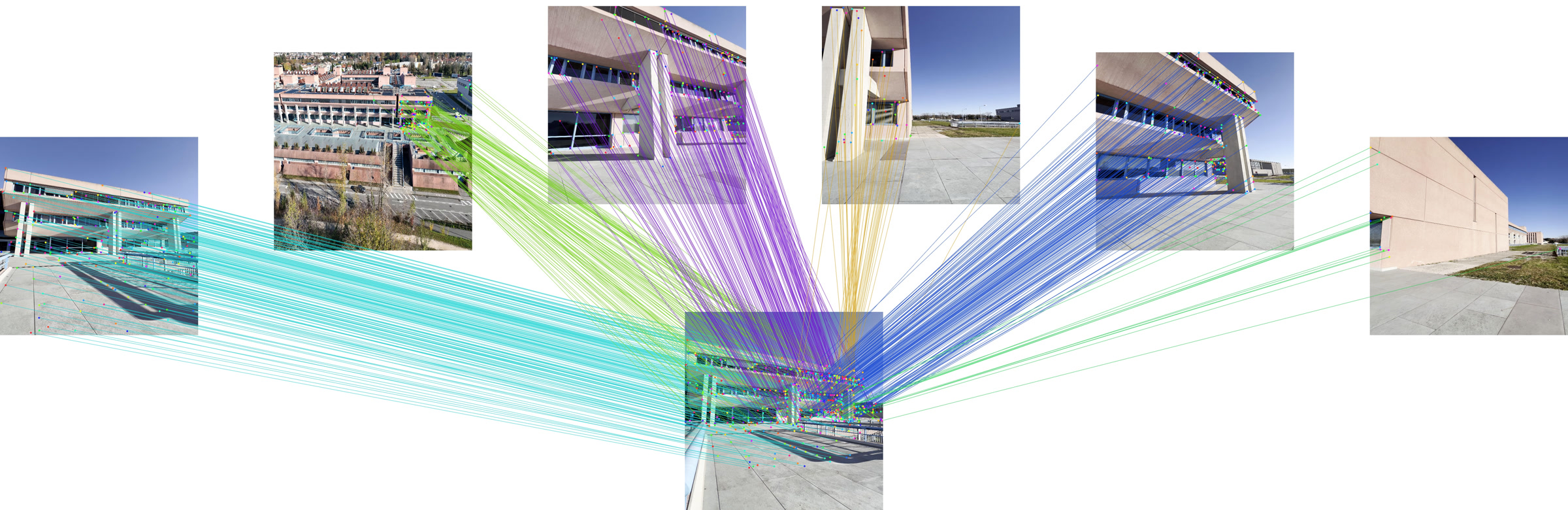}\end{minipage}%
    \hspace{\visright}}
\end{minipage}\par
\vfill
\par\addvspace{0.6em}\noindent
\begin{minipage}{\textwidth}
\makebox[\linewidth][c]{%
    \begin{minipage}[c]{0.12\textwidth}\centering\textbf{\scriptsize RGB input}\end{minipage}%
    \hspace{0.01\textwidth}%
    \begin{minipage}[c]{\viswidth}\includegraphics[width=\linewidth]{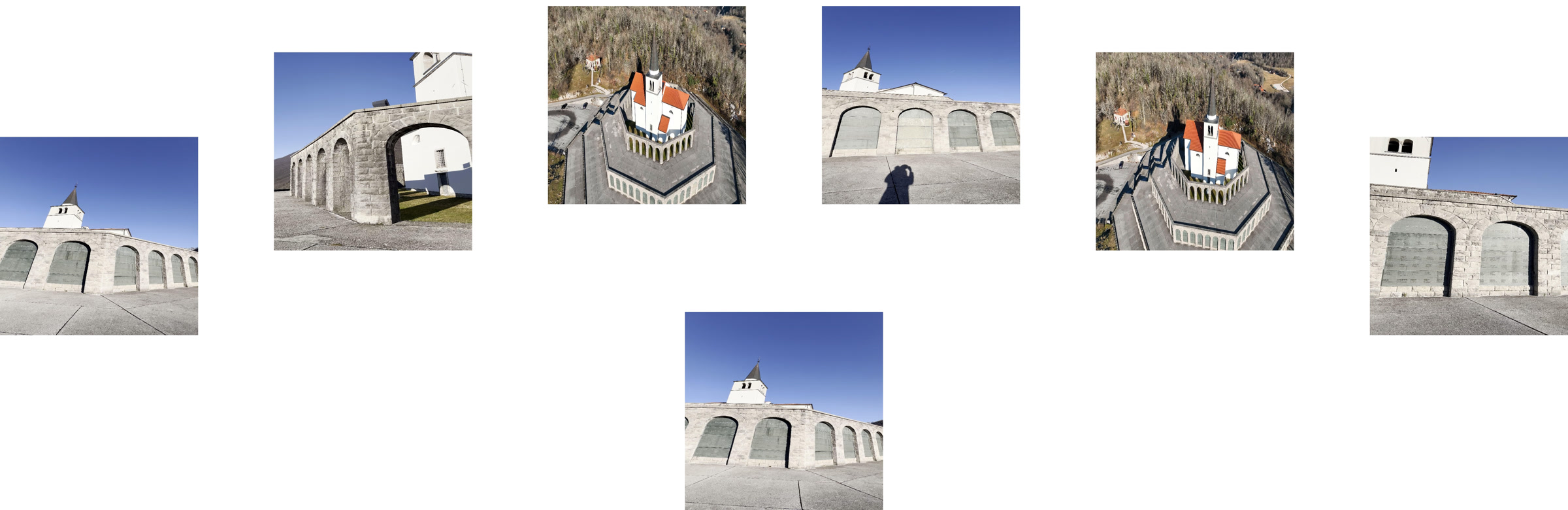}\end{minipage}%
    \hspace{\visright}}\par\vspace{-0.2em}
\makebox[\linewidth][c]{%
    \begin{minipage}[c]{0.12\textwidth}\centering\textbf{\scriptsize Star matching}\end{minipage}%
    \hspace{0.01\textwidth}%
    \begin{minipage}[c]{\viswidth}\includegraphics[width=\linewidth]{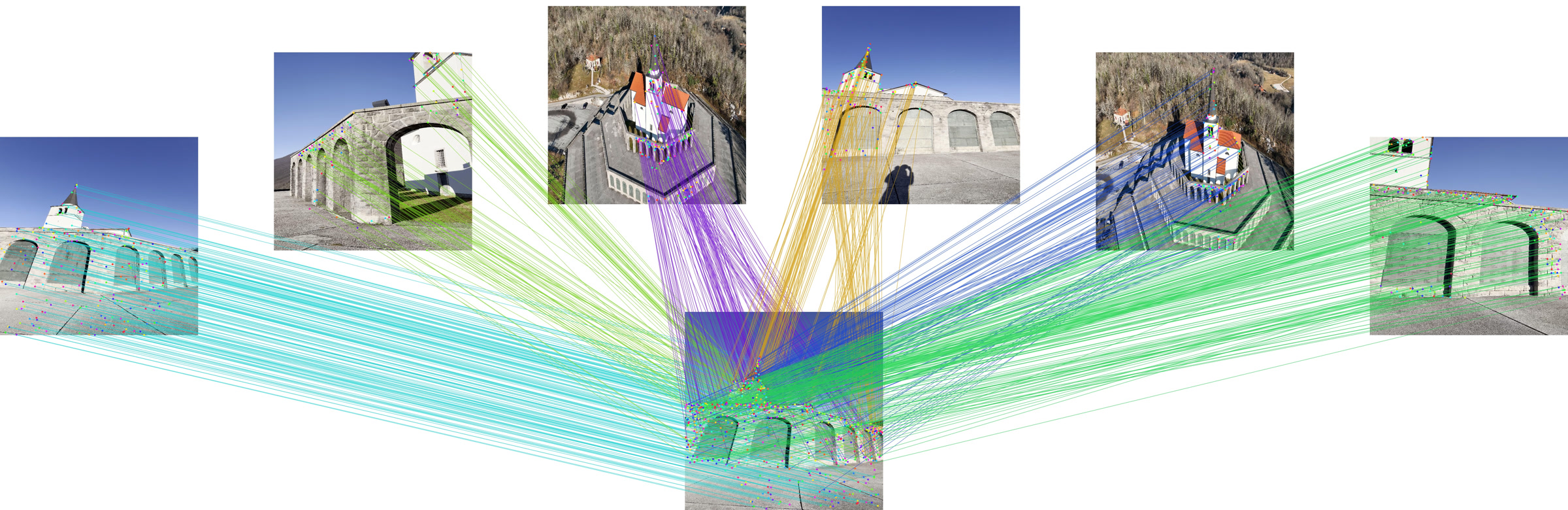}\end{minipage}%
    \hspace{\visright}}
\end{minipage}\par
\clearpage
\par\addvspace{0.6em}\noindent
\begin{minipage}{\textwidth}
\makebox[\linewidth][c]{%
    \begin{minipage}[c]{0.12\textwidth}\centering\textbf{\scriptsize RGB input}\end{minipage}%
    \hspace{0.01\textwidth}%
    \begin{minipage}[c]{\viswidth}\includegraphics[width=\linewidth]{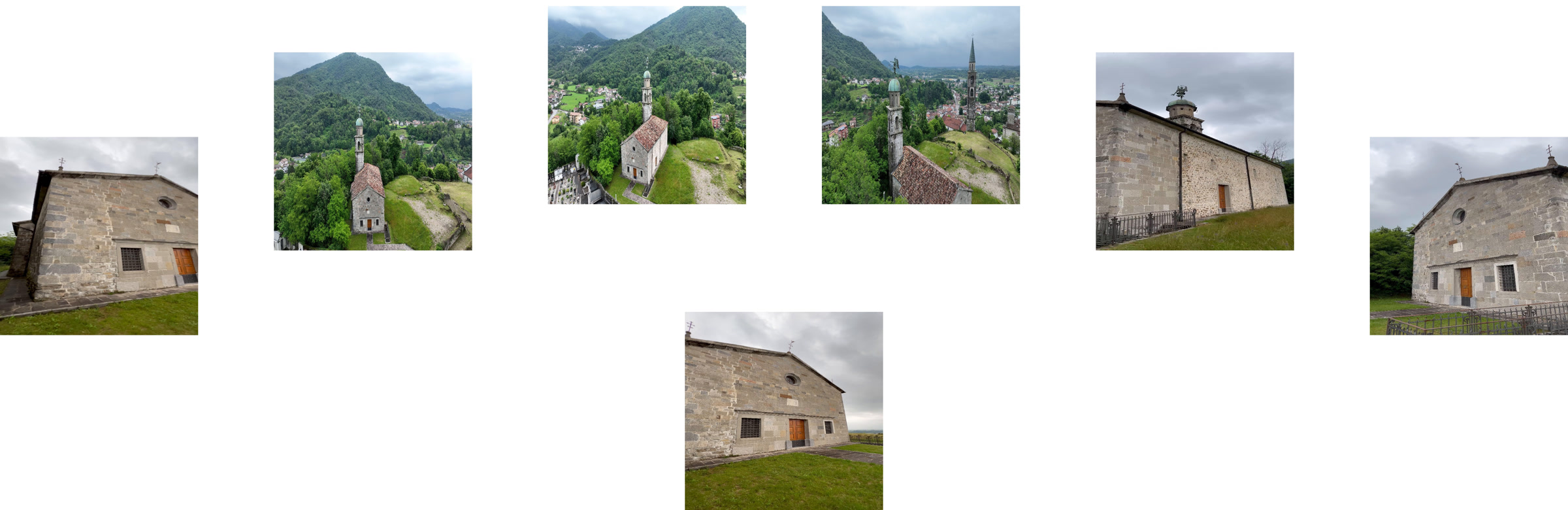}\end{minipage}%
    \hspace{\visright}}\par\vspace{-0.2em}
\makebox[\linewidth][c]{%
    \begin{minipage}[c]{0.12\textwidth}\centering\textbf{\scriptsize Star matching}\end{minipage}%
    \hspace{0.01\textwidth}%
    \begin{minipage}[c]{\viswidth}\includegraphics[width=\linewidth]{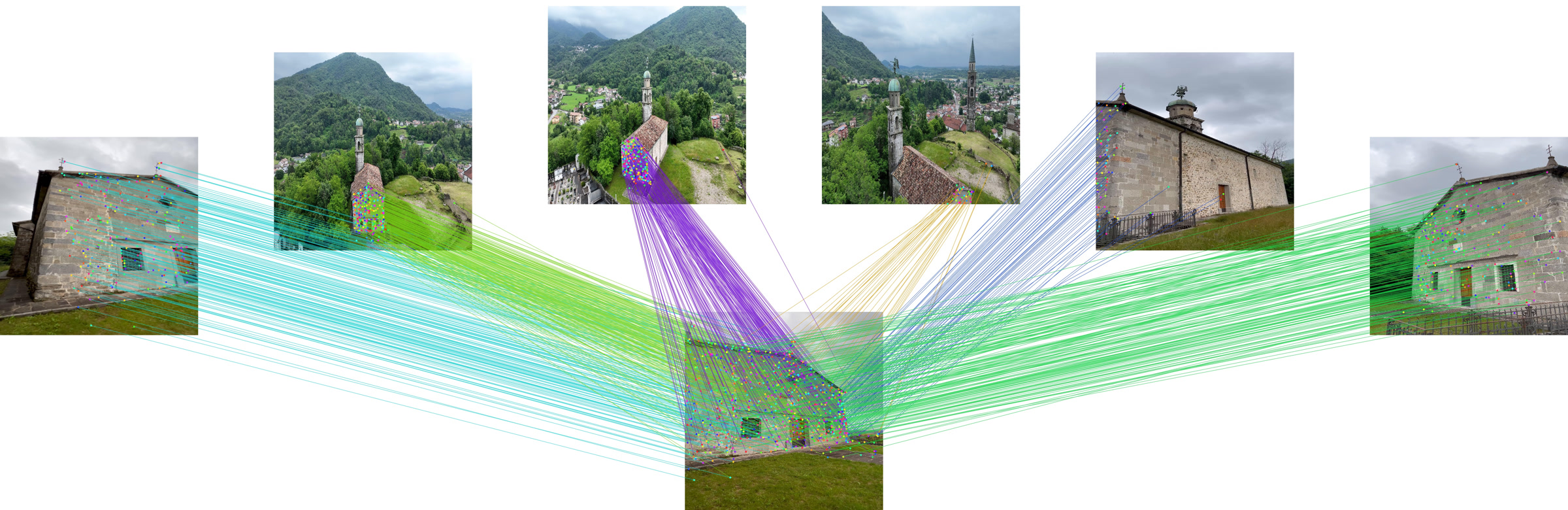}\end{minipage}%
    \hspace{\visright}}
\end{minipage}\par
\vspace{2.4em}
\par\addvspace{0.6em}\noindent
\begin{minipage}{\textwidth}
\makebox[\linewidth][c]{%
    \begin{minipage}[c]{0.12\textwidth}\centering\textbf{\scriptsize RGB input}\end{minipage}%
    \hspace{0.01\textwidth}%
    \begin{minipage}[c]{\viswidth}\includegraphics[width=\linewidth]{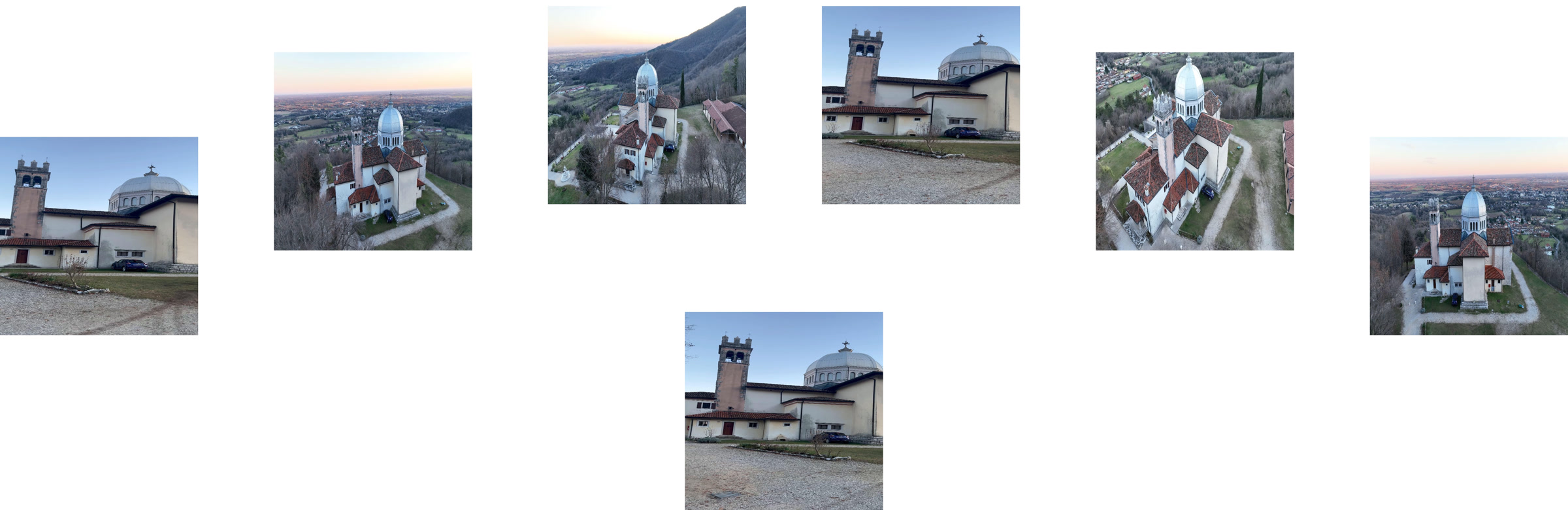}\end{minipage}%
    \hspace{\visright}}\par\vspace{-0.2em}
\makebox[\linewidth][c]{%
    \begin{minipage}[c]{0.12\textwidth}\centering\textbf{\scriptsize Star matching}\end{minipage}%
    \hspace{0.01\textwidth}%
    \begin{minipage}[c]{\viswidth}\includegraphics[width=\linewidth]{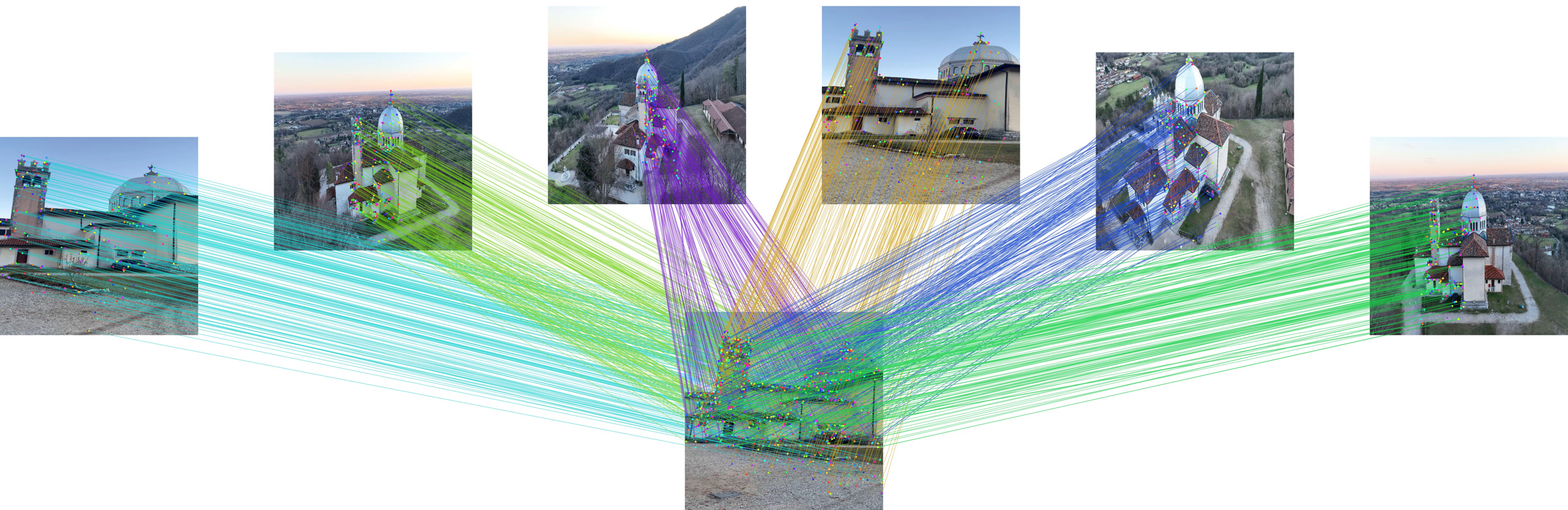}\end{minipage}%
    \hspace{\visright}}
\end{minipage}\par

\end{document}